%% file: acl_latex.tex
\pdfoutput=1

\documentclass[11pt]{article}

\usepackage[preprint]{acl}

\usepackage{times}
\usepackage{latexsym}
\usepackage[T1]{fontenc}

\usepackage[utf8]{inputenc}

\usepackage{microtype}

\usepackage{inconsolata}
\usepackage{amsfonts}

\usepackage{placeins}

\usepackage{graphicx}
\usepackage{amsmath}
\usepackage{float}  
\usepackage{booktabs}
\usepackage{multirow}
\usepackage{makecell}
\usepackage{float}
\usepackage{subcaption}
\usepackage{tcolorbox}
\usepackage{booktabs} 
\usepackage{colortbl} 
\usepackage[table]{xcolor} 
\usepackage[ruled,vlined]{algorithm2e} 
\usepackage{setspace}
\DontPrintSemicolon 
\usepackage{xurl}
\usepackage{afterpage}
\usepackage{amsmath,amssymb}

\usepackage{titlesec}

\titlespacing*{\paragraph}
  {0pt}              
  {0.6\baselineskip} 
  {0.5em}            
\title{Rethinking the Role of Efficient Attention in Hybrid Architectures}

\author{
Ziqing Qiao$^{1}$\thanks{Equal contribution.}, 
Yinuo Xu$^{1}$\footnotemark[1],
Chaojun Xiao$^{1}$\thanks{Corresponding authors},
Zhou Su$^{2}$,
Zihan Zhou$^{2}$,\\
{\bf Yingfa Chen}$^{1}$,
{\bf Xiaoyue Xu}$^{2}$,
{\bf Xu Han}$^{1}$\footnotemark[2],
{\bf Zhiyuan Liu}$^{1}$\footnotemark[2]\\
$^{1}$Tsinghua University $^{2}$OpenBMB \\
\texttt{qzq24@mails.tsinghua.edu.cn}, \texttt{\{xcj,han-xu,liuzy\}@tsinghua.edu.cn}
}

\begin{document}
\maketitle

\input{latex/Sec/0_abstract}

\input{latex/Sec/1_introduction}

\input{latex/Sec/2_relatedwork}
\input{latex/Sec/3_preliminary}
\input{latex/Sec/4_experiment}
\input{latex/Sec/5_analysis}
\input{latex/Sec/6_discussion}
\input{latex/Sec/7_conclusion}
\input{latex/Sec/8_limitation}

\bibliography{custom}
\input{latex/Sec/9_appendix}

\end{document}

%% file: latex/Sec/0_abstract.tex
\begin{abstract}
Modern language models increasingly adopt hybrid architectures that combine full attention with efficient attention modules, such as sliding-window attention (SWA) and recurrent sequence mixers. However, how these efficient modules shape model capabilities remains poorly understood. To address this gap, we conduct a systematic analysis across hybrid architectures from three perspectives: scaling behavior, mechanism analysis, and architecture design. First, from a scaling perspective, we find that efficient-attention design primarily affects how fast long-context capability emerges, while different hybrids eventually converge to comparable long-context performance under sufficient training. Second, mechanistically, we show that long-range retrieval is mainly carried by full attention, whereas efficient attention shapes its optimization trajectory. This explains a counter-intuitive phenomenon we call \emph{Large-Window Laziness}: larger SWA windows can delay the formation of retrieval heads in full-attention layers. Third, guided by this mechanism, we show that applying NoPE to only the full-attention layers of a small-window SWA hybrid substantially improves long-context performance with negligible impact on short-context performance.\footnote{We release our code at \href{https://github.com/thunlp/rethinking-hybrid-attention}{rethinking-hybrid-attention}.}
\end{abstract}

%% file: latex/Sec/1_introduction.tex
\section{Introduction}
\label{sec:intro}
As large language models are increasingly used for long-document understanding and agentic workflows, handling extended contexts has become a core requirement in recent model releases~\citep{deepseekai2026deepseekv4,singh2025openai}. However, standard softmax attention, which we refer to as \emph{full attention}, is costly at long sequence lengths~\citep{vaswani2017attention}. This has motivated a family of hybrid attention architectures that combine full attention with \emph{efficient attention} such as sliding-window attention (SWA)~\citep{beltagy2020longformer} and recurrent sequence mixers~\citep{gu2023mamba,yang2024gated}, a design now widely adopted in recent language models~\citep{agarwal2025gpt,gemma_2025,cao2026qwen3}.

Despite their prevalence, the role of efficient attention in hybrid architectures remains unclear. Existing studies lack a unified mechanistic analysis of how different efficient-attention designs shape the capabilities and training dynamics of hybrid architectures, particularly their long-context performance~\citep{xiao2026mimo,li2025minimax,wang2025systematic,bae2025hybrid}. To address this gap, we investigate three research questions:

\noindent
\emph{\textbf{RQ1 - Scaling Behavior}: How do hybrid architectures scale in short- and long-context performance?}

\noindent
\emph{\textbf{RQ2 - Mechanism Analysis}: How does efficient-attention design influence long-context performance?}

\noindent
\emph{\textbf{RQ3 - Architecture Design}: What design principles lead to more effective hybrid architectures?}

\paragraph{Scaling laws for short- and long-context capabilities.} We study how hybrid architectures scale in short- and long-context performance through the lens of \emph{scaling law} across multiple model scales and training budgets~\citep{kaplan2020scaling,hoffmann2022training}. 
Considering the discreteness and instability of downstream benchmark scores~\citep{liang2026revealing}, we use validation $\mathrm{Loss}$ and $\log(\mathrm{LongPPL})$~\citep{fang2024wrong} as two continuous fitting targets. The former captures general short-context modeling quality, while the latter provides a smooth proxy for long-context capability. 
The fitted scaling curves clearly show that efficient-attention design has little effect on validation $\mathrm{Loss}$, but leads to more pronounced differences in $\log(\mathrm{LongPPL})$. 
Specifically, different hybrid architectures exhibit substantial gaps under limited training budgets, with large-window SWA hybrids performing notably worse.
However, as training becomes more sufficient, these gaps shrink significantly and eventually approach a similar level.

\paragraph{Efficient attention as an optimization prior.}
The scaling pattern above leaves us with two seemingly contradictory puzzles. \textit{First}, why do hybrid architectures with different efficient attention ultimately converge to a similar long-context level? \textit{Second}, why do their convergence rates differ so much, especially across SWA variants with different window sizes? Our mechanistic analysis shows that both puzzles share a common explanation: efficient attention does not directly determine long-context capability; instead, it acts as an \emph{optimization prior} that shapes how full attention is trained.

\textit{Why do hybrids converge?} Through receptive-field constraint and layer-wise probing experiments, we find that long-range information is carried primarily by full attention rather than by efficient-attention modules, even recurrent sequence mixers with in-principle unbounded receptive fields. Sharing this same full-attention component, different hybrids converge to a similar long-context level regardless of their efficient-attention design.

\textit{Why do convergence rates differ?} While full attention sets the final converged level, efficient attention influences long-context capability by shaping how quickly full attention develops its long-range retrieval behavior during training. As a concrete example, by tracing retrieval heads~\citep{wu2025retrieval}, we find that retrieval heads form noticeably later in hybrid models equipped with larger SWA windows: once the local window already supplies sufficient context for next-token prediction, the gradient signal pushing full attention to learn long-range retrieval weakens. We term this phenomenon \emph{Large-Window Laziness}.

\paragraph{Hybrid architecture designs beyond efficient attention.}
These findings suggest that hybrid architecture design should focus less on increasing the intrinsic capability of efficient attention and more on helping full attention learn long-range retrieval more effectively. From this perspective, we revisit several design choices beyond the efficient-attention module. As a simple but effective instance, we apply NoPE~\citep{kazemnejad2023impact} to the full-attention layers of a small-window SWA hybrid. This simple modification yields a clear long-context capability gain with negligible impact on short-context performance, which is reflected consistently in downstream benchmark evaluations.

\begin{figure*}[htbp]
    \centering
    \includegraphics[width=1\linewidth]{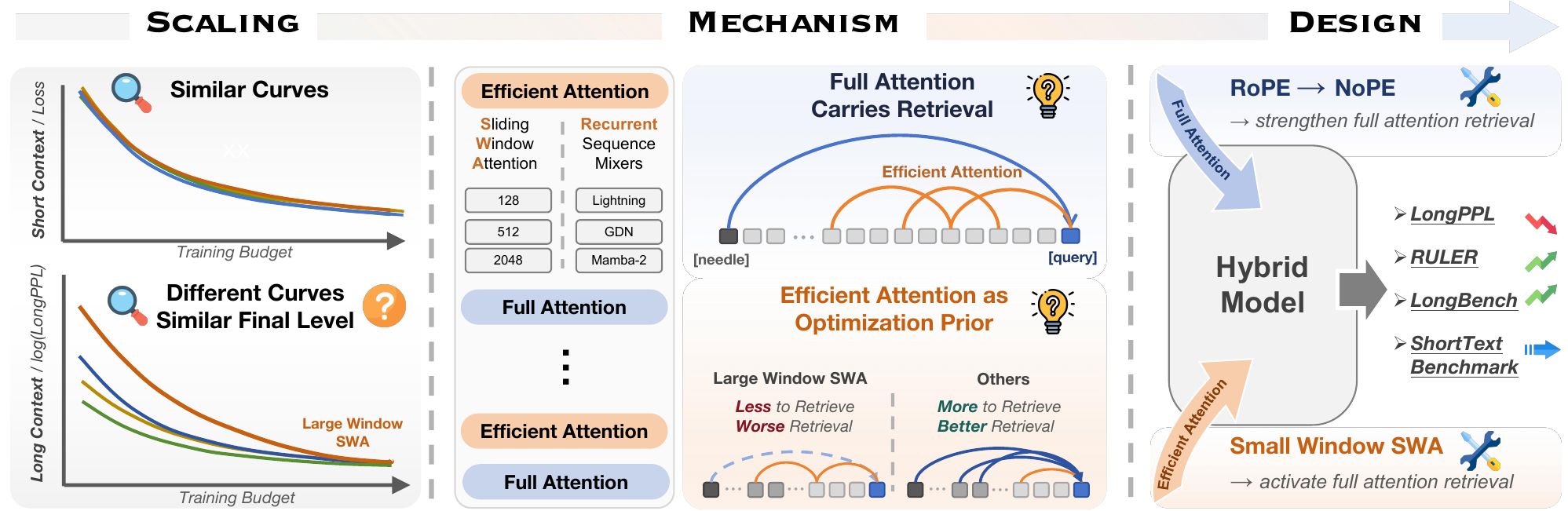}
    \caption{\textbf{Overview.}
    \textit{Scaling}: different efficient-attention designs yield distinct $\log(\mathrm{LongPPL})$ curves that converge to a similar level after sufficient training.
    \textit{Mechanism}: long-range retrieval is primarily carried by full attention, while efficient attention acts as an \emph{optimization prior}, where large-window SWA lags the most.
    \textit{Design}: strengthening full attention itself (\textit{e.g.}, RoPE$\rightarrow$NoPE in full attention) further improves long-context performance.}
    \label{fig:overview}
\end{figure*}

Figure~\ref{fig:overview} summarizes our main findings and their design implications. Taken together, our results reframe the role of efficient attention in hybrid architectures. The practical bottleneck for long-context capability is not simply how powerful the efficient-attention module is, but how it affects the emergence of long-range retrieval in full attention. This view explains the scaling patterns across hybrids and points to full attention as a key target for improving long-context hybrid models.

%% file: latex/Sec/2_relatedwork.tex
\section{Related Work}
\label{sec:related}

\paragraph{Hybrid Attention Architectures.}
Existing hybrid architectures mainly follow two lines. One uses SWA~\citep{beltagy2020longformer} as efficient attention, where recent designs have moved toward smaller windows and sparser full-attention ratios with limited overall performance degradation~\citep{agarwal2025gpt,huang2026step35flashopen}. The other employs recurrent sequence mixers that compress past history into a compact recurrent state, such as Lightning Attention~\citep{qin2024various}, Mamba-2~\citep{dao2024transformers}, and Gated DeltaNet~\citep{yang2025gated}, which are increasingly adopted in recent models~\citep{li2025minimax,blakeman2025nvidia,cao2026qwen3,team2026minicpm}. Beyond the choice of efficient-attention module, recent work also explores head-wise mixing~\citep{dong2025hymba,xiao2025wunenghybridstateattention} and positional encoding for the full-attention layers~\citep{yang2026rope,puvvada2025swan,chen2026hybrid}. However, most of these studies present only final results or limited ablations within specific systems~\citep{gemma_2025,xiao2026mimo}, leaving a lack of controlled comparisons across efficient-attention architectures.

Several studies have begun to examine structural choices in hybrid architectures more systematically. \citet{wang2025systematic} compare multiple linear attention variants and mixing ratios, while \citet{waleffe2024empirical,bae2025hybrid} analyze layer composition and placement in Mamba-Transformer hybrids. Yet these studies remain within recurrent-mixer-based hybrids and lack a mechanistic explanation. We bridge this gap by comparing different efficient-attention designs under a controlled scaling-law setup and analyzing how they shape the long-context capability of hybrid architectures.

\paragraph{Scaling Laws and Long-Context Evaluation.}
Scaling laws characterize how pretraining performance depends on model and data scale~\citep{kaplan2020scaling,hoffmann2022training}, with subsequent extensions to transfer learning~\citep{hernandez2021scaling} and downstream capability prediction~\citep{chen2024scaling}. However, scaling laws for long-context capability remain underexplored. Existing long-context evaluations typically rely on discrete benchmarks such as RULER and LongBench~\citep{hsieh2024ruler,bai2024longbench}, which measure final performance but are less suitable for tracking pretraining dynamics. A complementary line of mechanistic studies shows that retrieval heads underlie long-context factual recall~\citep{wu2025retrieval,xiao2025duoattention} and tracks the formation of retrieval heads to observe how long-context capability develops during pretraining~\citep{liang2026revealing}, but such signals describe the mechanism rather than quantify capability. In contrast, LongPPL~\citep{fang2024wrong} provides a continuous perplexity-style metric that correlates strongly with long-context benchmarks, and has since been adopted in recent long-context studies~\citep{song2026towards,willette2026delta}. We further leverage this metric to fit scaling laws for long-context performance, enabling a more comprehensive comparison of how long-context capability emerges across hybrid architectures.

%% file: latex/Sec/3_preliminary.tex
\section{Preliminaries}
\label{sec:pre}


\subsection{Hybrid Architecture}
\label{sec:pre:hybrid}

We cover two common forms of efficient attention: \textbf{Sliding-Window Attention (SWA)}, where each token attends only to a finite local window, and \textbf{recurrent sequence mixers}, including \textbf{Lightning Attention}, \textbf{Mamba-2}, and \textbf{Gated DeltaNet (GDN)}, which compress past tokens into a recurrent state through different decay strategies and update rules.

We use $q_t,k_t,v_t\in\mathbb{R}^{d_h}$ for the per-head query, key, and value vectors at position $t$ (with $d_k{=}d_v{=}d_h$ assumed for notational simplicity), and let $\mathrm{softmax}_s$ denote the softmax normalized over the index $s$. The formulas below present canonical forms of the mechanisms; implementation-level parameter choices used for matching sizes of different hybrid models are given in Appendix~\ref{app:models}.

\paragraph{Full Attention.}
For each position $t$, the output $O_t$ is computed over all preceding positions $s\le t$ :
\begin{equation}
  O_t = \sum_{s\le t}\mathrm{softmax}_s\!\bigl(q_t^{\!\top} k_s/\sqrt{d_h}\bigr)\,v_s
  \label{eq:softmax-attn}
\end{equation}

\paragraph{Sliding Window Attention.}
SWA restricts the summation range to a window of size $w$:
\begin{equation}
  O_t = \sum_{s\in[t{-}w{+}1,\,t]}\mathrm{softmax}_s\!\bigl(q_t^{\!\top} k_s/\sqrt{d_h}\bigr)\,v_s
  \label{eq:swa-attn}
\end{equation}

The three recurrent mixers below all share the form $O_t = S_t q_t$ with a recurrent state $S_t\in\mathbb{R}^{d_h\times d_h}$; they differ mainly in how $S_t$ is updated.

\paragraph{Lightning Attention.} Lightning is a linear attention with a fixed per-head decay $\gamma\in(0,1)$:
\begin{equation}
  S_t = \gamma S_{t-1} + v_t k_t^{\!\top}.
  \label{eq:lightning}
\end{equation}

\paragraph{Mamba-2.} Following the structured state-space duality (SSD) form, Mamba-2 can be written as:
\begin{equation}
  S_t = \gamma_t S_{t-1} + v_t k_t^{\!\top}.
  \label{eq:mamba-2}
\end{equation}
The data-dependent $\gamma_t$ allows per-token control over how much of the past state is preserved.

\paragraph{Gated DeltaNet.} GDN further adds controlled forgetting through a data-dependent decay $\alpha_t \in (0,1)$ and a data-dependent update strength $\beta_t \in (0,1)$:
\begin{equation}
\begin{aligned}
  S_t &= \alpha_t S_{t-1}(I-\beta_t k_t k_t^{\!\top})
       + \beta_t v_t k_t^{\!\top}.
\end{aligned}
\label{eq:gdn}
\end{equation}
Here, the delta-rule term removes the existing content associated with $k_t$ before writing the new association $v_t k_t^{\!\top}$.

\begin{figure*}[htbp]
    \centering
    \includegraphics[width=1\linewidth]{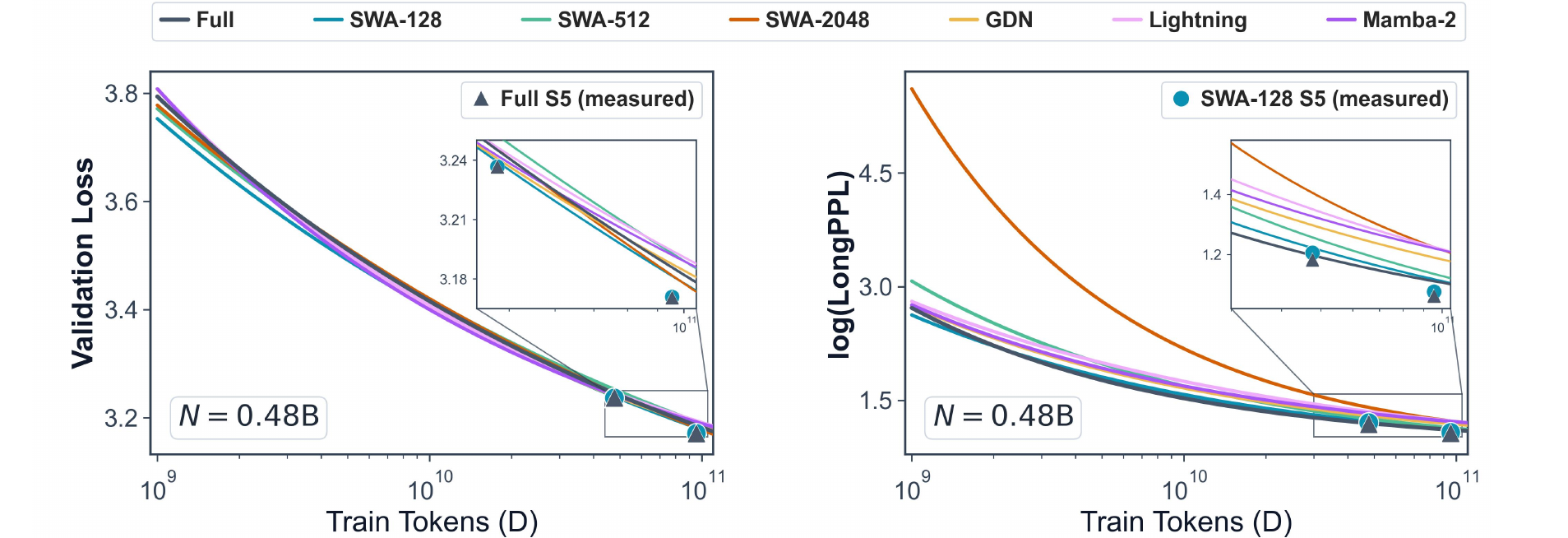}
        \caption{\textbf{Predicted $\mathrm{Loss}$ and $\log(\mathrm{LongPPL})$ at S5 scale ($N{=}0.48\mathrm{B}$) across Train tokens $D$.} $\mathrm{Loss}$ curves of all hybrids closely overlap, whereas $\log(\mathrm{LongPPL})$ curves show large gaps in the low-data regime that shrink with more training. The insets verify extrapolation accuracy against the measured S5 checkpoints of \emph{Full} and \emph{SWA-128}.}
    \label{fig:compare}
\end{figure*}

\subsection{Scaling Law}

To compare hybrid architectures across model scales and training budgets, we use two fitting targets: validation $\mathrm{Loss}$ for short-context modeling and $\log(\mathrm{LongPPL})$ for long-context capability.

\paragraph{Loss.}
Validation loss is the standard target in language-model scaling laws \citep{kaplan2020scaling,hoffmann2022training}. We select $40$K held-out samples from C4 \citep{raffel2020exploring} that are disjoint from our training corpus, and report the average negative log-likelihood as $\mathrm{Loss}$. 

\paragraph{LongPPL.}



We adopt $\log(\mathrm{LongPPL})$ \citep{fang2024wrong} as the fitting target for long-context capability. Following its original implementation, we adopt GovReport \citep{huang2021efficient} as the evaluation corpus and Llama-3.1-8B \citep{grattafiori2024llama} as the reference model. More details are provided in Appendix~\ref{app:longppl}.

\paragraph{Scaling Law Formula.}
For both $\log(\mathrm{LongPPL})$ and $\mathrm{Loss}$, we model performance as a function of model parameters $N$ (w/o embeddings) and training tokens $D$ \citep{hoffmann2022training}, using the separable power-law form as the fitting template: 
\begin{equation}
  L(N, D) = a N^{-\alpha} + b D^{-\beta}
  \label{eq:scaling-law}
\end{equation}
where $a,b,\alpha,\beta$ are fitted separately for each architecture and fitting target.

%% file: latex/Sec/4_experiment.tex
\section{Scaling Behavior of Short- and Long-Context Capabilities}
\label{sec:exp}

To answer RQ1, we fit scaling laws for validation $\mathrm{Loss}$ and $\log(\mathrm{LongPPL})$ to compare short-context and long-context capabilities across hybrid architectures with different efficient-attention designs.
 
\subsection{Settings}
\label{exp:set}

\paragraph{Model architecture.}
We compare a full-attention Transformer baseline, denoted as \emph{Full}, with six layer-wise hybrid architectures that differ in efficient-attention components. Three hybrids use SWA with window sizes of 128, 512, and 2048, denoted as \emph{SWA-128}, \emph{SWA-512}, and \emph{SWA-2048}. The other three use recurrent sequence mixers, denoted as \emph{Lightning}, \emph{Mamba-2}, and \emph{GDN}. All hybrid models alternate full-attention and efficient-attention layers with a $1{:}1$ ratio.

\begin{table}[t]
    \centering
    \caption{Key hyperparameters of \emph{Full} model for S1--S5.}
    \label{tab:hyperparams}
    \resizebox{\columnwidth}{!}{
        \begin{tabular}{lccccc}
            \toprule
            \textbf{Configuration} & \textbf{S1} & \textbf{S2} & \textbf{S3} & \textbf{S4} & \textbf{S5} \\
            \midrule
            Params (w/o embed.) & 15M & 31M & 65M & 104M & 477M\\
            Total Params & 71M & 107M & 159M & 217M & 665M \\
            Layers & 10 & 12 & 16 & 18 & 30 \\
            Hidden dim & 384 & 512 & 640 & 768 & 1280 \\
            FFN dim & 960 & 1280 & 1600 & 1920 & 3200 \\
            Heads (Q) & 6  & 8  & 10  & 12  & 20  \\
            Heads (KV) & 2  & 2  & 2  & 2  & 2  \\
            Head dim & 64 & 64 & 64 & 64 & 64 \\
            \bottomrule
        \end{tabular}
    }
\end{table}


\paragraph{Scaling setup.}
The scaling study covers five model sizes, S1--S5, with the hyperparameters of the \emph{Full} configuration summarized in Table~\ref{tab:hyperparams}. For the main scaling analysis, we evaluate S1--S4 checkpoints trained with six token budgets, $D \in \{100N, 200N, 300N, 400N, 500N, 1000N\}$, across all architectures, where $N$ corresponds to the parameters without embedding. For S5 scale ($N=0.48B$ excluding embeddings; total parameters $0.66B$), we train \emph{Full} and \emph{SWA-128} at $D=100N$ and $D=200N$ for larger-scale extrapolation checks.

All models are pretrained with a 16K context length on a $1{:}1$ mixture of long and short datasets, which allows us to simultaneously measure short- and long-context capabilities. More training settings are given in Appendix~\ref{app:training}.

\subsection{Scaling Law of Validation $\mathbf{Loss}$}
\label{sec:exp:loss-scaling}
We fit the scaling law for validation $\mathrm{Loss}$ using 18 data points from S1--S3, and hold out the 6 data points from S4 as a verification set. As shown in Figure~\ref{fig:loss_scaling_full}, all seven architectures are well captured by the scaling law, achieving high $R^2$ on both the fitting and verification sets.

To compare architectures under matched scaling conditions, we examine the predicted $\mathrm{Loss}$ at the S5 scale ($N{=}0.48\mathrm{B}$) across training tokens $D$, and include the measured S5 $\mathrm{Loss}$ to assess the extrapolation accuracy of the fitted curves.

As shown in the left panel of Figure~\ref{fig:compare}, the validation $\mathrm{Loss}$ curves of all hybrid models closely overlap with \emph{Full} across the full range of $D$. This indicates that efficient-attention design has limited impact on short-context capability.

\subsection{Scaling Law of $\log(\mathbf{LongPPL})$}
\label{sec:exp:longppl-scaling}

We fit the scaling law for $\log(\mathrm{LongPPL})$ following the same protocol as for validation $\mathrm{Loss}$, except that we exclude the S1 checkpoint at $D=100N$ because its training budget is too small for stable long-context behavior. As shown in Figure~\ref{fig:longppl_scaling_full}, although $\log(\mathrm{LongPPL})$ is noisier at early checkpoints, it is still smoothly captured by Eq.~\eqref{eq:scaling-law}.

In contrast to the strong overlap observed for $\mathrm{Loss}$, the predicted $\log(\mathrm{LongPPL})$ reveals much larger architectural differences. We compare architectures under the same setting as above and include the measured S5 $\log(\mathrm{LongPPL})$ values to assess extrapolation accuracy.

As shown in the right panel of Figure~\ref{fig:compare}, a clear pattern emerges: architectural differences are most pronounced in early training, corresponding to the low-data regime, where large-window SWA, especially \emph{SWA-2048}, exhibits substantially higher $\log(\mathrm{LongPPL})$. As the training becomes more sufficient, this gap rapidly shrinks, and the hybrid models with different efficient-attention designs eventually converge to similar levels with \emph{Full}.

Taken together, the $\mathrm{Loss}$ and $\log(\mathrm{LongPPL})$ scaling results reveal a clear separation between final capability and training dynamics: \textbf{Efficient-attention design has limited effect on the eventual short- and long-context capabilities of hybrid models, but strongly shapes the emergence speed of long-context capability.} 

%% file: latex/Sec/5_analysis.tex
\section{Mechanism: How Efficient Attention Shapes Long-Context Capability}
\label{sec:analysis}
The key observation in Section~\ref{sec:exp} naturally motivates \emph{\textbf{RQ2:} How does efficient-attention design influence long-context performance?} In this section, we conduct a series of mechanistic experiments that dissect the role of efficient attention in long-context modeling; full implementation details and extended analyses can be found in Appendix~\ref{app:mechanism}.

\subsection{The Dominant Role of Full Attention}
\label{sec:analysis:ga-dominates}
A natural hypothesis is that efficient attention with a larger receptive field, especially recurrent sequence mixers whose receptive field is in principle unbounded, should help improve the long-context capability of hybrid models. However, this is not supported by the scaling pattern that different hybrid models converge to similar $\log(\mathrm{LongPPL})$. To examine where long-context capability actually arises, we conduct a receptive-field constraint and a layer-wise probing experiment.

\begin{figure}[t]
    \centering
    \includegraphics[width=1\linewidth]{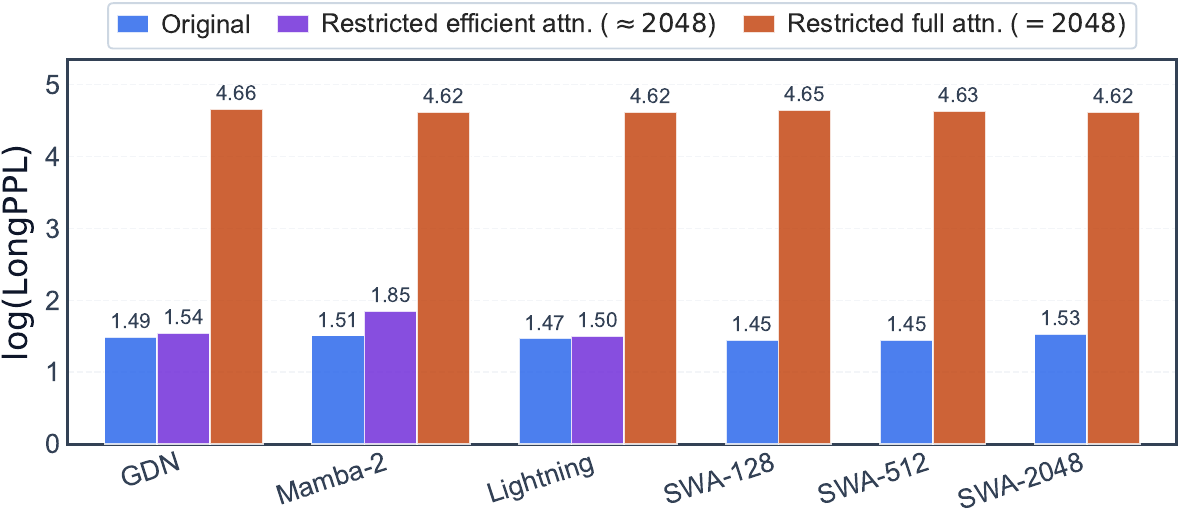}
    \caption{\textbf{Inference-time receptive-field restriction for S4/$1000N$ hybrids.} Restricting efficient attention to $\approx 2048$ tokens leaves $\log(\mathrm{LongPPL})$ nearly unchanged, while restricting full attention raises it sharply.}
    \label{fig:horizon}
\end{figure}

\paragraph{Receptive-field constraint.}
For the S4 models trained with $D=1000N$ in scaling experiments, we separately restrict the accessible receptive field of efficient attention and full attention to $\approx 2048$ tokens at inference time, and measure the resulting change in $\log(\mathrm{LongPPL})$. As shown in Figure~\ref{fig:horizon}, when the receptive field of full attention is restricted, $\log(\mathrm{LongPPL})$ increases sharply across all hybrid models. In contrast, restricting the receptive field of efficient attention causes only minor changes. This indicates that, in our setting, even recurrent sequence mixers whose receptive field is in principle unbounded and whose update rules are delicate, such as GDN, store little long-range information in their recurrent states during inference.

\paragraph{Probing Experiment.}
To examine how long-range information emerges across layers, we conduct a layer-wise probing experiment \citep{belinkov2022probing} on a Needle-in-a-Haystack (NIAH) \citep{hsieh2024ruler} classification task. For each layer, we extract the hidden state of the final query token and train a logistic-regression classifier to predict the inserted needle. By comparing the incremental change in probing accuracy from one layer to its predecessor, we estimate how much long-range information is introduced by each layer. Details are provided in Appendix~\ref{app:probing}.

\begin{figure}[t]
    \centering
    \includegraphics[width=\linewidth]{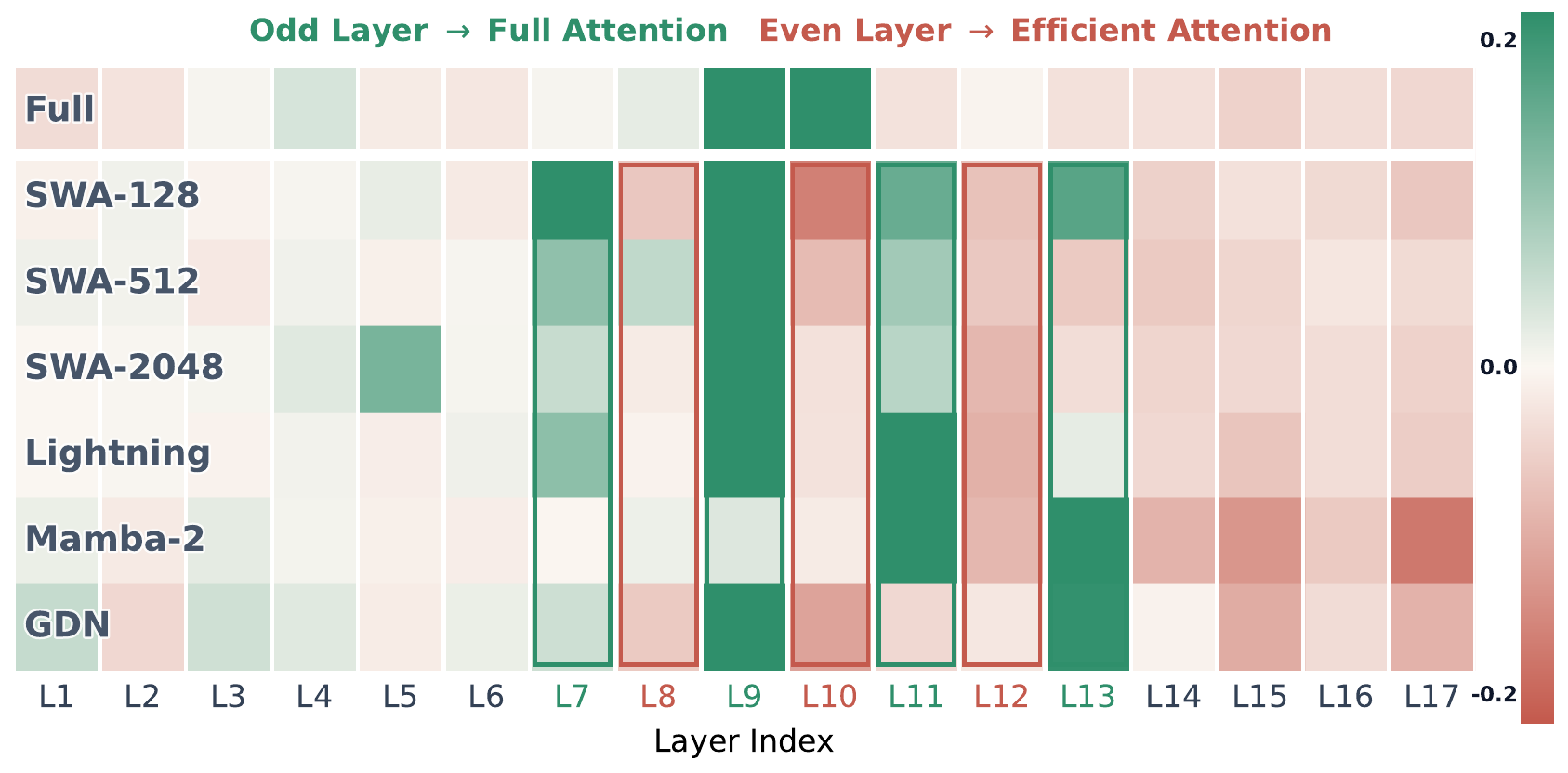}
    \caption{\textbf{Layer-wise probing accuracy gain on NIAH for the S4/$1000N$ models.} Cells show incremental accuracy over the previous layer. In all hybrids, gains concentrate at middle full-attention layers (odd-numbered).}
    \label{fig:probing}
\end{figure}

\begin{figure*}[t]
    \centering
    \begin{subfigure}[t]{0.32\linewidth}
        \centering
        \includegraphics[width=\linewidth]{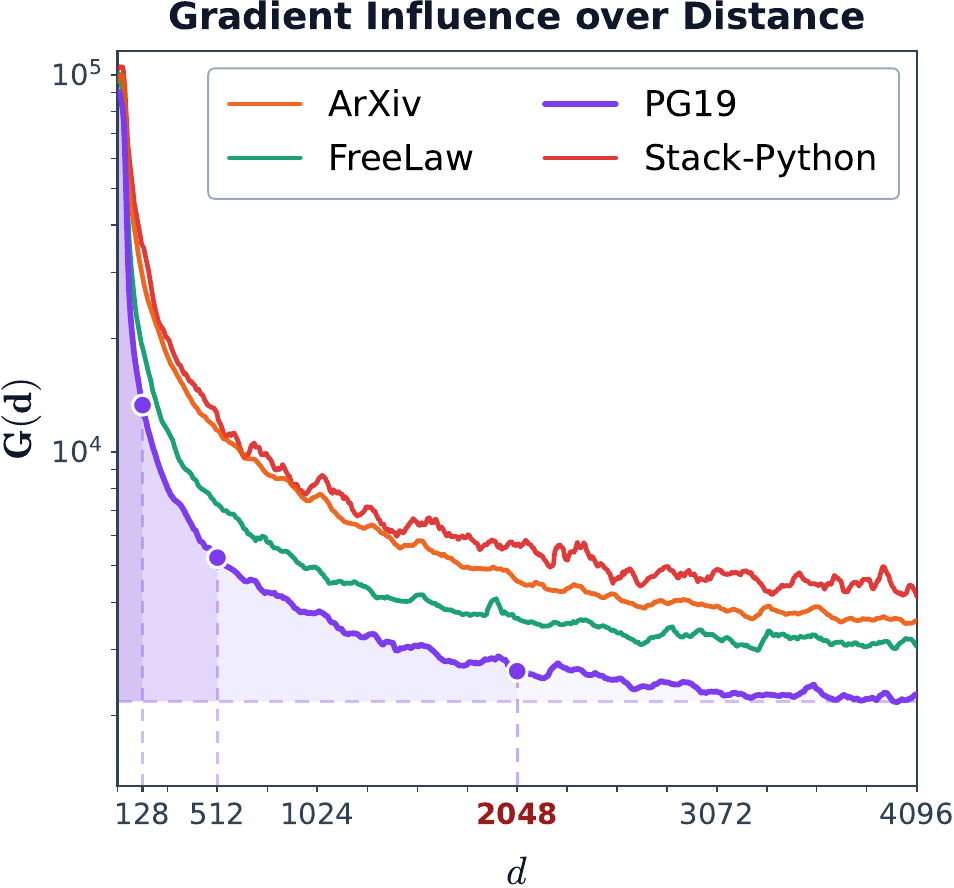}
        \caption{Gradient influence over distance.}
        \label{fig:gradient_influence}
    \end{subfigure}
    \hfill
    \begin{subfigure}[t]{0.64\linewidth}
        \centering
        \includegraphics[width=\linewidth]{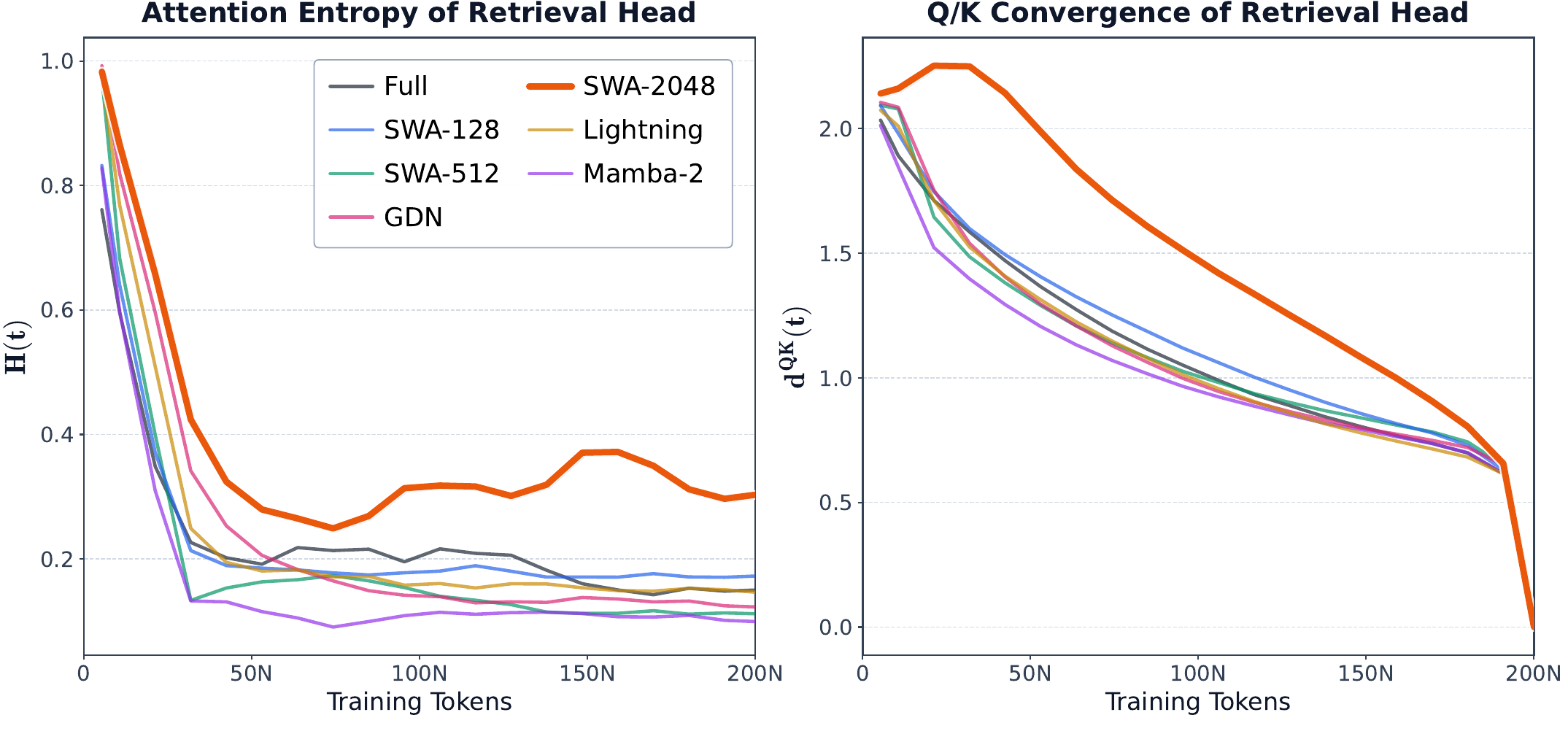}
        \caption{Retrieval-head training trajectories.}
        \label{fig:tracing}
    \end{subfigure}
    \caption{\textbf{Evidence for Large-Window Laziness.} \textbf{(a)} Beyond $2048$ tokens, $G(d)$ decays to a flat baseline, while the 512--2048 range still carries substantial signal. \textbf{(b)} \emph{SWA-2048} is the outlier: its retrieval-head attention entropy $H(t)$ stays high and Q/K weight distance $d^{\mathrm{QK}}(t)$ shrinks more slowly, indicating under-trained retrieval.}
    \label{fig:laziness_evidence}
\end{figure*}

Figure~\ref{fig:probing} shows that, in layer-wise hybrids, probing accuracy increases almost exclusively at middle full-attention layers \emph{(odd-numbered)}, while middle efficient-attention layers \emph{(even-numbered)} contribute little gain and even reduce accuracy. In contrast, \emph{Full} shows continuous growth across middle layers. This supports the view that long-range information in hybrids is mainly introduced and processed by full attention.

The receptive-field constraint and probing experiments suggest that long-context capability in hybrid architectures primarily relies on full attention rather than efficient-attention modules. This also helps explain the scaling behavior observed in Section~\ref{sec:exp:longppl-scaling}: architectural gaps in $\log(\mathrm{LongPPL})$ shrink after sufficient training because all hybrid models share the same full-attention design as \emph{Full}.

\subsection{Efficient Attention as an Optimization Prior of Long-Context Capability}
\label{sec:analysis:prior}

The scaling experiments show that different efficient-attention designs substantially affect the convergence speed of $\log(\mathrm{LongPPL})$. Since long-context capability is primarily carried by full attention, we argue that these differences arise because efficient attention affects how fast full attention learns long-range retrieval. This effect is especially clear in large-window SWA hybrids, which we refer to as \emph{Large-Window Laziness}.

Concretely, a large local window can already cover many useful dependencies during training. As a result, the model can often predict the next token using information within the sliding window, without relying on full attention to retrieve from farther positions. This weakens the optimization pressure for full attention to develop long-range retrieval ability, causing this ability to emerge more slowly. In contrast, SWA with smaller windows leaves more dependencies outside the window range, forcing the model to access them through full attention and thereby providing a denser signal for long-range retrieval. We provide two pieces of evidence consistent with this mechanism.

\paragraph{Gradient Influence Profiling.}
To estimate how next-token-prediction signal decays with distance $d$, we use Llama-3.1-8B to measure the gradient influence $G(d)$ \citep{li2016visualizing} on long documents sampled from the pretraining corpus. This proxy assumes that the natural long-range dependency distribution is largely model-agnostic, so the profile approximates the dependency signal seen during hybrid-model pretraining. For an input sequence $x_{1:T}$, we define $G(d)$ as
\[
G(d)
=
\mathbb{E}_{x}
\left[
\left\|
\frac{\partial s(x)}
{\partial e_{T-d}}
\right\|_2
\right],
\]
where $e_{T-d}$ is the embedding of the token at distance $d$, and $s(x)$ denotes the logit used for prediction. This quantity measures how sensitive the model's prediction is to each historical token, and thus serves as a proxy for distance-dependent signal strength. As shown in Figure~\ref{fig:gradient_influence}, the signal beyond 2048 tokens decays to a flat baseline, while the 512--2048 range still contains substantial gradient signal. This suggests that a 2048-token window already captures most useful training signal, whereas sub-512 windows leave substantial signal outside the window, thereby imposing stronger pressure on full attention to learn retrieval. This is consistent with \emph{Large-Window Laziness}.

\paragraph{Retrieval-Head Tracing.}
We use retrieval heads \citep{wu2025retrieval} as the unit of analysis: we densely save intermediate checkpoints before the S4 models reach $D=200N$ tokens, identify retrieval heads in the final checkpoint, and track two diagnostics at each intermediate checkpoint $t$.

(i) $H(t)$, the normalized attention entropy when retrieving the needle token in the NIAH task:
\[
H(t)
=
-\frac{1}{\log |\mathcal{V}_q|}
\sum_{j \in \mathcal{V}_q}
a^{(t)}_{qj}\log a^{(t)}_{qj},
\]
where $a^{(t)}_{qj}$ is the attention weight from query $q$ to visible key $j$ at checkpoint $t$, and $\mathcal{V}_q$ is the visible-key set. Lower $H(t)$ indicates sharper retrieval.

(ii) $d^{\mathrm{QK}}(t)$, the relative parameter distance from checkpoint $t$ to the final checkpoint:
\[
d^{\mathrm{QK}}(t)
=
\sum_{W \in \{W_Q, W_K\}}
\frac{\|W^{(t)}-W^{(t_{end})}\|_F}{\|W^{(t_{end})}\|_F},
\]
where $W_Q$ and $W_K$ are the query and key projection matrices of the identified retrieval head, $\|\cdot\|_F$ denotes the Frobenius norm, and $t_{end}$ indexes the $D=200N$ checkpoint. We report the mean of both diagnostics over the Top-2 retrieval heads.

Figure~\ref{fig:tracing} shows that \emph{SWA-2048} follows a clearly different pattern from the other models: its normalized attention entropy remains high, and its retrieval-head weights converge more slowly, indicating that its retrieval heads remain under-trained.
By contrast, retrieval heads train faster in smaller-window SWA and recurrent efficient-attention hybrids, consistent with the need for full attention to access information beyond what the efficient-attention module can reliably provide. We provide additional analyses of retrieval-head formation from complementary perspectives in Appendix~\ref{sec:retrieval}, all of which lead to consistent conclusions.

Together, these analyses yield a unified mechanistic answer to \emph{\textbf{RQ2}}: \textbf{efficient attention primarily shapes how efficiently full attention learns long-range retrieval, rather than carrying long-range information directly.} 

%% file: latex/Sec/6_discussion.tex
\section{Hybrid Architecture Design Beyond Efficient Attention}
\label{sec:discussion}

\begin{table*}[!t]
  \centering
  \caption{\textbf{Downstream evaluation} of \emph{Full}, \emph{SWA-128}, and \emph{SWA-128-NoPE} at S4 ($0.22B$) and S5 ($0.66B$). RULER\textsubscript{NIAH} is the average over the 8 NIAH-style sub-tasks in RULER; ShortAvg is the average over 19 short-context benchmarks, evaluated with the 16K models. \textbf{Bold} marks the best within each model scale. Full per-task results are reported in Appendix~\ref{sec:benchmark_evaluation}.}
  \label{tab:downstream_verification}
  \small
  \setlength{\tabcolsep}{5pt}
  \resizebox{\textwidth}{!}{%
    \begin{tabular}{llc|ccc|ccc}
      \toprule
      \multirow{2}{*}{Setting} & \multirow{2}{*}{Model}
      & \multirow{2}{*}{ShortAvg}
      & \multicolumn{3}{c|}{Long-Context (16K)}
      & \multicolumn{3}{c}{Long-Context (32K)} \\
      \cmidrule(lr){4-6} \cmidrule(lr){7-9}
      & & & RULER & RULER\textsubscript{NIAH} & LongBench
      & RULER & RULER\textsubscript{NIAH} & LongBench \\
      \midrule
      \multirow{3}{*}{\makecell{$S4 (0.22B)$\\$D\approx 100B$}}
      & Full
      & \textbf{38.13}
      & 25.09 & 35.95 & 15.09
      & -- & -- & -- \\
      & SWA-128
      & 38.03
      & 35.33 & 49.58 & 15.88
      & -- & -- & -- \\
      & SWA-128-NoPE
      & 37.88
      & \textbf{44.80} & \textbf{67.81} & \textbf{16.43}
      & -- & -- & -- \\
      \midrule
      \multirow{3}{*}{\makecell{$S5 (0.66B)$\\$D \approx 100B$}}
      & Full
      & 40.46
      & 47.17 & 67.14 & 18.44
      & 43.90 & 62.61 & 18.93 \\
      & SWA-128
      & 41.31
      & 46.13 & 65.91 & 17.52
      & 41.86 & 60.17 & 18.30 \\
      & SWA-128-NoPE
      & \textbf{41.32}
      & \textbf{52.88} & \textbf{82.31} & \textbf{19.02}
      & \textbf{46.98} & \textbf{70.42} & \textbf{19.46} \\
      \bottomrule
    \end{tabular}
  }
\end{table*}
The mechanism above motivates us to revisit hybrid architecture design, raising \emph{\textbf{RQ3:} What design principles lead to more effective hybrid architectures?} We move beyond efficient attention and examine several other design factors through scaling law and downstream benchmark evaluation.

\subsection{Full-to-Efficient Layer Ratio}

\begin{figure}[t]
  \centering
  \includegraphics[width=1\linewidth]{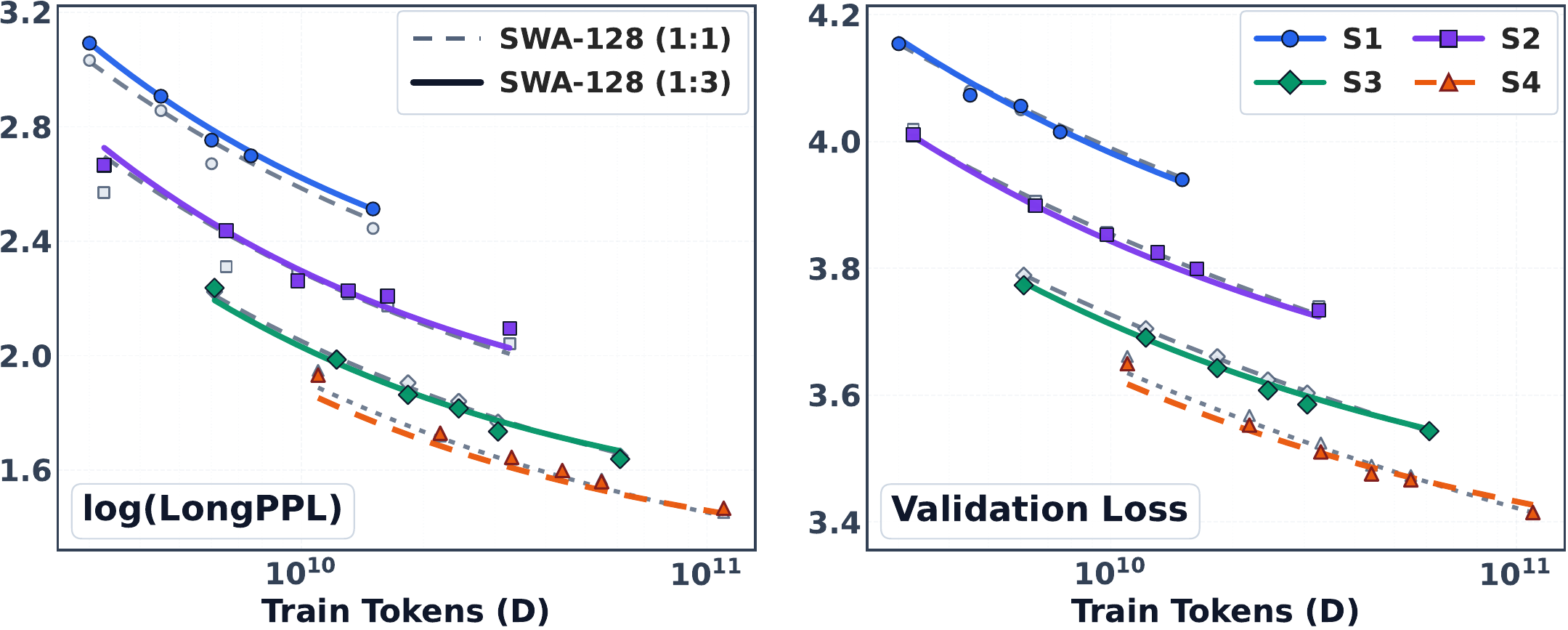}
  \caption{SWA-128 (1:1) vs. SWA-128 (1:3).}
  \label{fig:ratio_compare}
\end{figure}

\begin{figure}[t]
  \centering
  \includegraphics[width=1\linewidth]{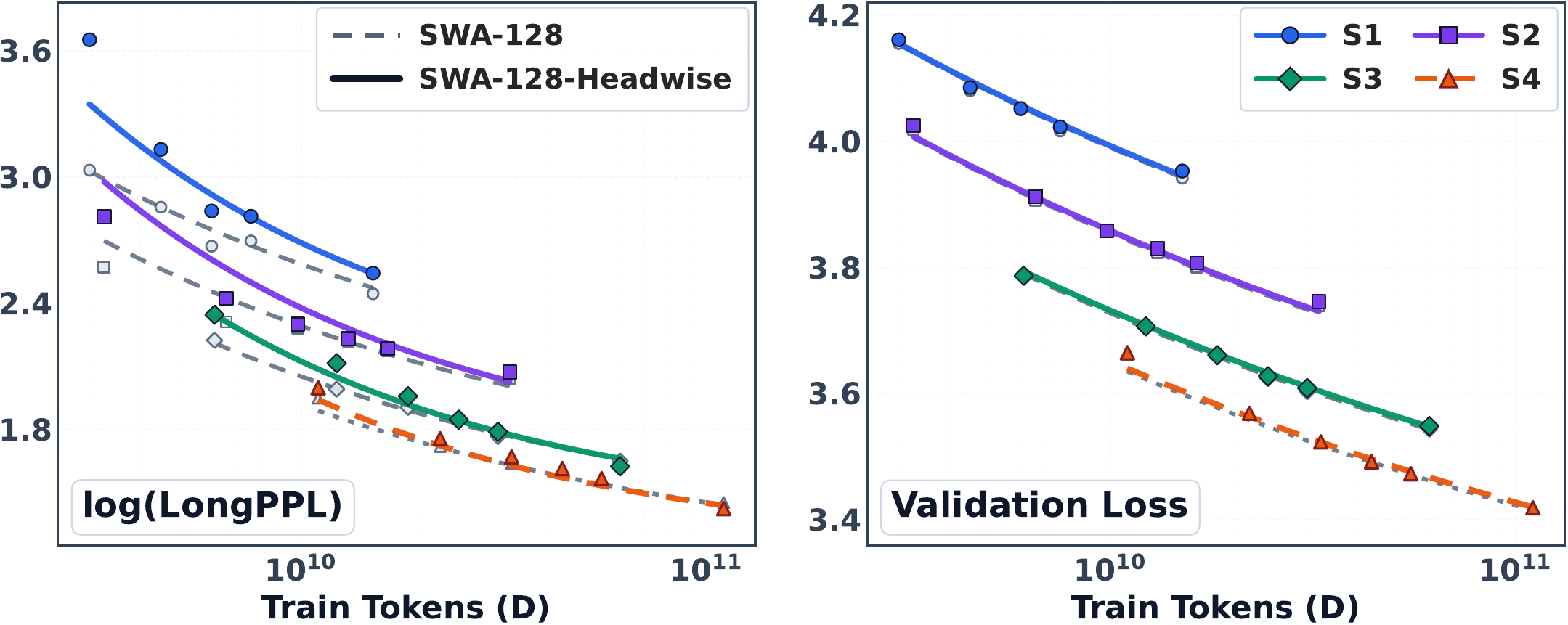}
  \caption{SWA-128 vs. SWA-128-Headwise.}
  \label{fig:head_compare}
\end{figure}

\begin{figure}[t]
  \centering
  \includegraphics[width=1\linewidth]{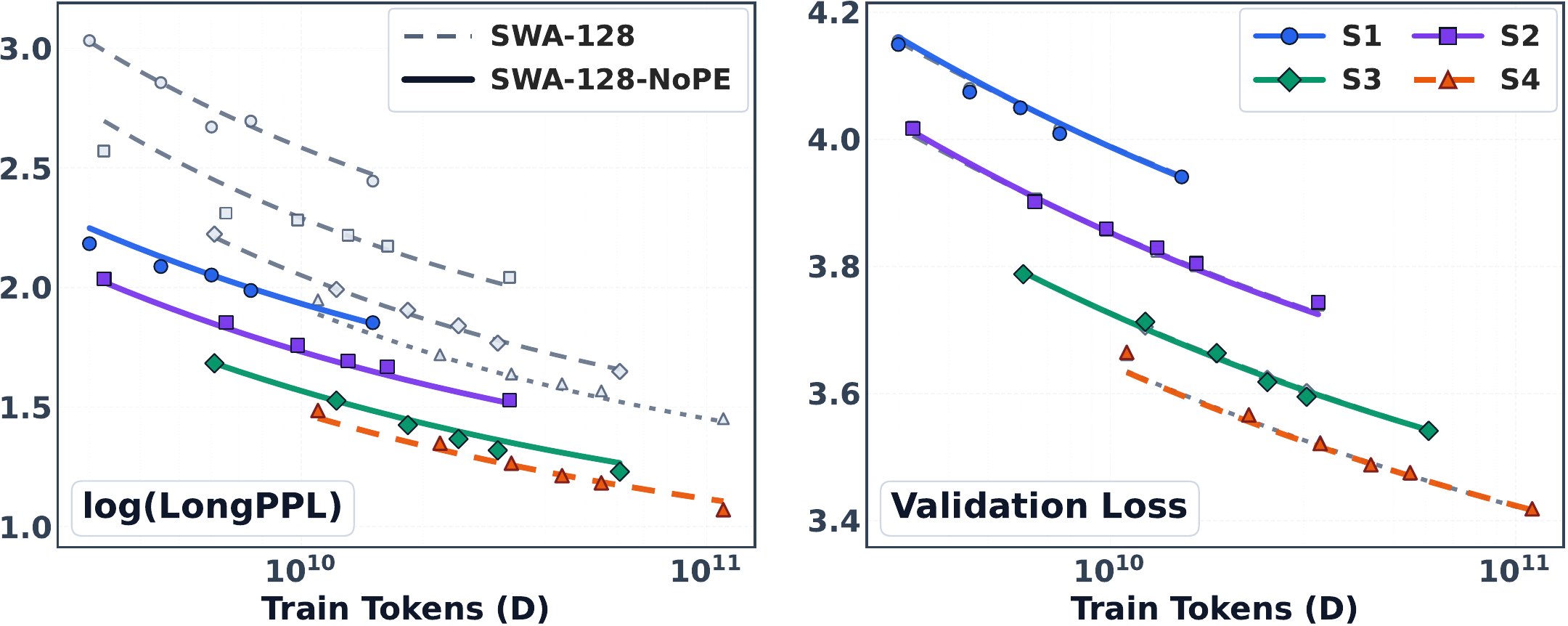}
  \caption{SWA-128 vs. SWA-128-NoPE.}
  \label{fig:nope_compare}
\end{figure}

We compare the 1:1 SWA-128 setting used in our main experiments with a sparser 1:3 variant, and fit their validation $\mathrm{Loss}$ and $\log(\mathrm{LongPPL})$ scaling curves. As shown in Figure~\ref{fig:ratio_compare}, the 1:3 ratio gives almost the same validation $\mathrm{Loss}$ as the 1:1 ratio. For $\log(\mathrm{LongPPL})$, however, the sparser model performs worse at small scales, likely because the number of full-attention layers is too limited. As model size increases, this gap closes, suggesting that full-attention density can be safely reduced once enough full-attention layers are available.
\subsection{Layer-wise vs. Head-wise}
Another design choice is whether to place full attention in dedicated layers or distribute it across heads within each layer, as in recent head-wise or intra-layer hybrid designs \citep{bae2025hybrid}. To examine this factor, we compare the layer-wise SWA-128 model with a head-wise variant, SWA-128-Headwise. As shown in Figure~\ref{fig:head_compare}, under our setting, head-wise mixing does not provide an advantage over layer-wise. Specifically, the two methods reach similar validation $\mathrm{Loss}$ and $\log(\mathrm{LongPPL})$ after sufficient training, yet the head-wise variant shows slower $\log(\mathrm{LongPPL})$ convergence.

\subsection{Positional Encoding of Full Attention}
\label{sec:discussion:nope}
Recent studies show that applying NoPE to full-attention layers can effectively enhance their long-range retrieval capability \citep{yang2026rope,puvvada2025swan}. We take \emph{SWA-128} as the base since it activates full-attention retrieval well, and apply NoPE to its full-attention layers, denoted as \emph{SWA-128-NoPE}. As shown in Figure~\ref{fig:nope_compare}, this change substantially decreases $\log(\mathrm{LongPPL})$ while leaving validation $\mathrm{Loss}$ nearly unchanged.

Following the training protocol of scaling experiments, we train \emph{Full}, \emph{SWA-128}, and \emph{SWA-128-NoPE} at S4 ($0.22\mathrm{B}$) and S5 ($0.66\mathrm{B}$) under ${\approx}100\mathrm{B}$ tokens. To further evaluate in longer contexts, we continue training the S5 checkpoints for an additional $5\mathrm{B}$ tokens at a 32K sequence length. For long-context, we use RULER \citep{hsieh2024ruler} and LongBench \citep{bai2024longbench}; for short-context, we report the average over 19 benchmarks. As shown in Table~\ref{tab:downstream_verification}, \emph{SWA-128-NoPE} consistently leads on long-context benchmarks at both scales while remaining comparable on short-context tasks.


The design studies suggest that hybrid architecture design should move beyond simply choosing a stronger efficient-attention component and instead prioritize choices that \textbf{better activate or directly strengthen full attention, allowing its long-range retrieval capability to emerge more efficiently.}

%% file: latex/Sec/7_conclusion.tex
\section{Conclusion}
Through scaling-law fits and mechanistic analysis, we find that the long-context performance of hybrid models is primarily determined by full attention, while efficient attention, acting as an \emph{optimization prior}, indirectly shapes it by modulating how quickly full attention learns long-range retrieval. This suggests that, under limited training budgets, hybrid design should favor choices that more effectively activate and strengthen the long-context capability of full attention, such as small-window SWA and NoPE, both validated in our experiments.

%% file: latex/Sec/8_limitation.tex
\section*{Limitations}

Although our experiments cover multiple model scales and verify the fitted scaling laws via extrapolation, the largest model we train is still at the sub-billion-parameter level with at most $\approx\!100$B pretraining tokens, which is smaller than the scale of frontier industrial systems. We also pretrain directly at a 16K context length and extend to at most 32K, in contrast to the prevailing recipe that pretrains on short context first and subsequently extends to long context. These choices may limit the applicability of our conclusions to larger-scale or differently trained settings.

For efficient-attention designs, we cover representative operators widely adopted in recent hybrid architectures, while leaving out some other popular variants such as RWKV-7~\citep{peng2025rwkv7gooseexpressivedynamic} and Kimi-Linear~\citep{team2025kimi}. In addition, the design choices discussed in Section~\ref{sec:discussion} are intended to validate our mechanistic conclusions rather than to serve as a full design study, and a more comprehensive verification at larger scales is left to future work.

%% file: latex/Sec/9_appendix.tex
\cleardoublepage
\appendix
\begin{figure*}[!htp]
    \centering
    \includegraphics[width=1\linewidth]{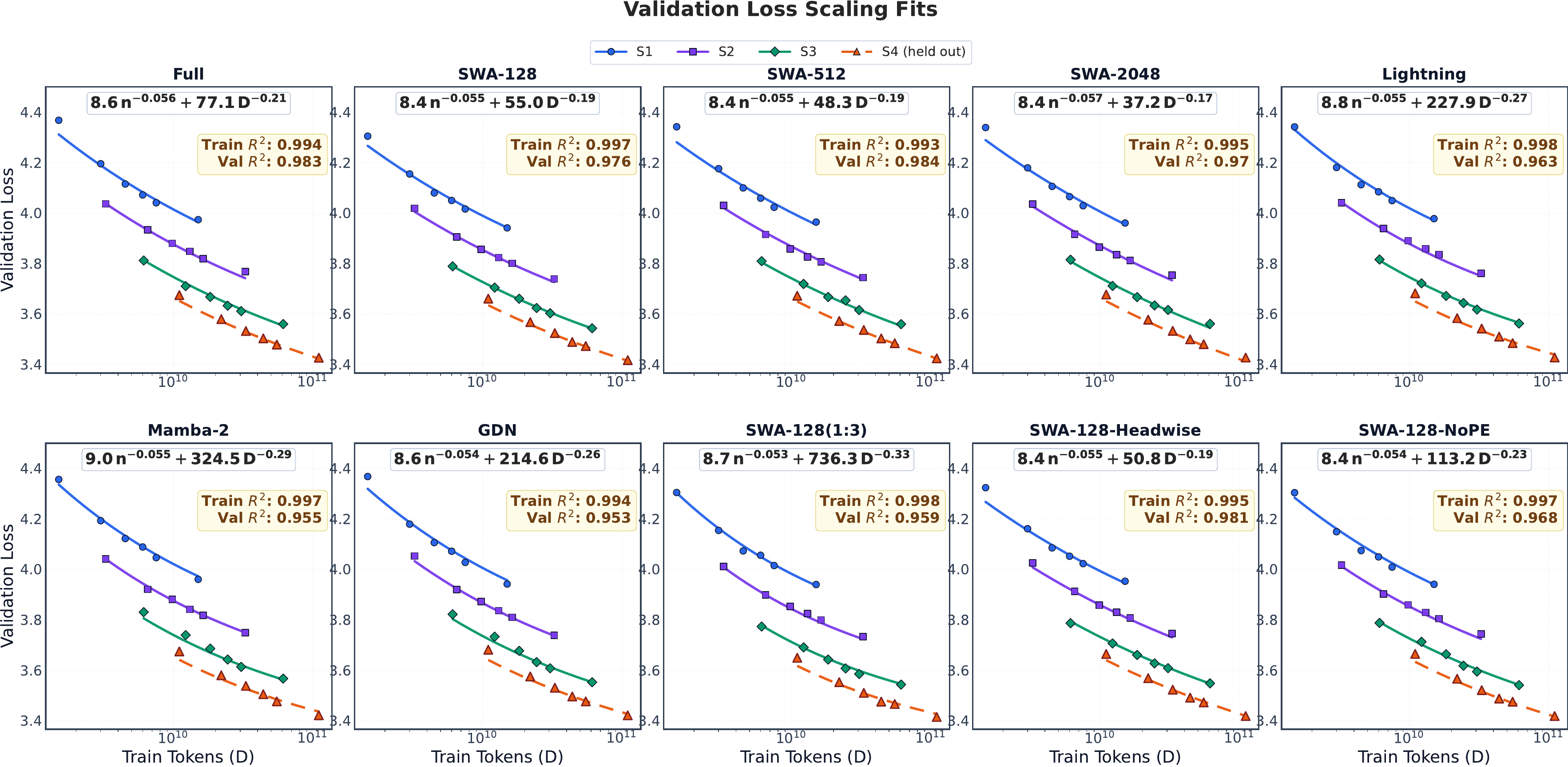}
    \caption{\textbf{Validation $\mathrm{Loss}$ scaling-law fits across all ten architectures.} Each panel plots validation $\mathrm{Loss}$ against training tokens $D$ for one architecture at four model scales S1--S4, with the 18 S1--S3 points used for fitting (solid markers) and the 6 S4 points held out for verification (orange triangles). Each colored curve is the fit of $L(N,D) = a N^{-\alpha} + b D^{-\beta}$ (Eq.~\eqref{eq:scaling-law}) at the corresponding $N$, with the fitted coefficients and the train/verification $R^2$ printed inside each panel. The first seven panels cover the architectures studied in the main scaling experiments (Section~\ref{sec:exp})---\emph{Full} together with three SWA hybrids and three recurrent-mixer hybrids---while the last three panels (\emph{SWA-128(1:3)}, \emph{SWA-128-Headwise}, and \emph{SWA-128-NoPE}) correspond to the design variants from Section~\ref{sec:discussion}. }
    \label{fig:loss_scaling_full}
\end{figure*}

\begin{figure*}[!t]
    \centering
    \includegraphics[width=1\linewidth]{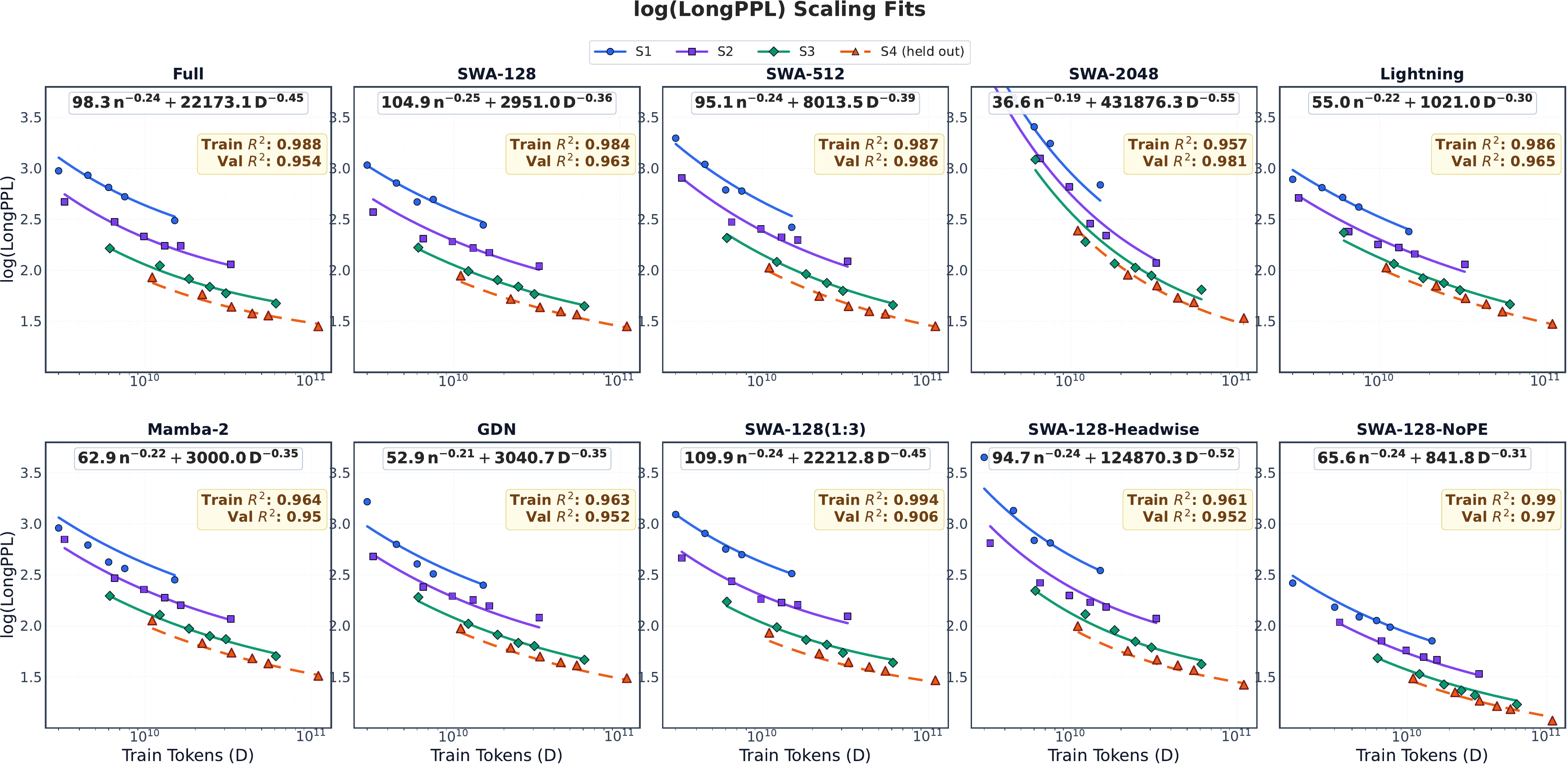}
    \caption{\textbf{$\log(\mathrm{LongPPL})$ scaling-law fits across the same ten architectures.} The panel layout, marker convention, and per-panel annotations follow Figure~\ref{fig:loss_scaling_full}. Compared with validation $\mathrm{Loss}$, $\log(\mathrm{LongPPL})$ is noticeably noisier at early checkpoints; in particular, the S1/$D=100N$ checkpoint is excluded from fitting due to unstable long-context behavior at this small training budget, leaving 17 S1--S3 fitting points and 6 S4 held-out points per architecture. Despite the higher noise level, the same power-law form $L(N,D) = a N^{-\alpha} + b D^{-\beta}$ still fits well. }
    \label{fig:longppl_scaling_full}
\end{figure*}

\section{LongPPL Evaluation Details}
\label{app:longppl}
LongPPL evaluates a model only on tokens whose prediction benefits from long context. Following \citet{fang2024wrong}, we identify these tokens by comparing the token-level negative log-likelihoods assigned by a reference model under full context and under a local chunk. In our experiments, we use GovReport \citep{huang2021efficient} as the evaluation corpus and Llama-3.1-8B \citep{grattafiori2024llama} as the reference model.

Let $\ell^{\mathrm{full}}_{\mathrm{ref}}(x_t)$ and $\ell^{\mathrm{chunk}}_{\mathrm{ref}}(x_t)$ denote the token-level negative log-likelihoods assigned by the reference model to token $x_t$ under the full context $x_{<t}$ and under a local chunk, respectively. The set of key tokens is defined as
\begin{equation}
  \begin{aligned}
  \mathcal{K} = \bigl\{\, t \,:\,
    &\ell^{\mathrm{chunk}}_{\mathrm{ref}}(x_t)
    - \ell^{\mathrm{full}}_{\mathrm{ref}}(x_t) > \tau_{\mathrm{gain}}, \\
    &\ell^{\mathrm{full}}_{\mathrm{ref}}(x_t) < \tau_{\mathrm{nll}}
    \,\bigr\},
  \quad \tau_{\mathrm{gain}} = \tau_{\mathrm{nll}} = 2 .
  \end{aligned}
  \label{eq:app-key-token}
\end{equation}
The first condition selects tokens that receive a clear gain from long context, while the second filters out tokens that remain hard to predict even with full context. For a model $M$ under evaluation, LongPPL is then computed only over $\mathcal{K}$:
\begin{equation}
  \mathrm{LongPPL}(M) = \exp\!\Bigl(
    \frac{1}{|\mathcal{K}|} \sum_{t\in\mathcal{K}}
    \ell^{\mathrm{full}}_{M}(x_t) \Bigr).
  \label{eq:app-longppl}
\end{equation}

\paragraph{Evaluation dataset statistics.}
Table~\ref{tab:eval_dataset_stats} summarizes the datasets used for the two scaling-law targets. The C4 validation split contains many short documents, making it suitable for measuring short-context modeling quality, whereas the GovReport subset used by LongPPL contains substantially longer sequences and a sufficient number of reference-selected key tokens per example. For GovReport, token lengths are computed after re-tokenization with the evaluated-model tokenizer and truncation at 16K, which is the pretraining sequence length of our models.

In preliminary experiments, we found that examples with fewer than 10 key tokens often produce unstable LongPPL estimates that occasionally spike to extremely large values. We therefore skip these examples to obtain more stable LongPPL estimates.

\begin{table*}[t]
    \centering
    \setlength{\tabcolsep}{5pt}
    \begin{tabular}{lccccc}
        \toprule
        Dataset / metric & Samples & After filter & Avg tokens & Median tokens & Avg key tokens \\
        \midrule
        C4 validation / $\mathrm{Loss}$ & 40{,}000 & 40{,}000 & 497.6 & 269 & -- \\
        GovReport / LongPPL & 10{,}000 & 8{,}898 & 13{,}317.3 & 13{,}276 & 78 \\
        \bottomrule
    \end{tabular}
    \caption{Evaluation dataset statistics. Token lengths use the evaluated-model tokenizer; GovReport lengths are after re-tokenization and truncation at 16K. Key tokens are identified by the Llama-3.1-8B reference; ``After filter'' is the count remaining after dropping examples with fewer than 10 key tokens (no filter is applied to C4).}
    \label{tab:eval_dataset_stats}
\end{table*}

\section{Model Details}
\label{app:models}
To compare hybrid architectures fairly, we keep the backbone configuration of \emph{Full}, \emph{SWA}, \emph{Lightning}, \emph{Mamba-2}, and \emph{GDN} matched as closely as possible, including the number of layers, hidden size, GQA grouping, and per-head dimension. For efficient attention variants that introduce additional parameters, we make only minimal architectural adjustments so that the total parameter count stays close to the \emph{Full} backbone. This avoids mixing the benefit of extra modules from the original implementations into our comparison of efficient-attention designs.

\subsection{Softmax Attention}
For softmax attention, prior work has observed the attention-sink phenomenon, where attention probability can concentrate on a small number of non-semantic positions at the beginning of the sequence \citep{xiao2024efficient}. To mitigate this, we adopt a learnable per-head softmax sink, as used in recent open models \citep{agarwal2025gpt}. Concretely, for head $h$ the attention distribution is
\[
    a_{ij}^{(h)}
    =
    \frac{
        \exp\!\left(q_i^{(h)\top}k_j^{(h)} / \sqrt{d_h}\right)
    }{
        \exp(s_h)
        \;+\;
        \sum_{\ell \le i}
        \exp\!\left(q_i^{(h)\top}k_\ell^{(h)} / \sqrt{d_h}\right)
    },
\]
where $s_h$ is a learnable per-head scalar initialized to zero. This is equivalent to introducing a virtual ``sink'' key with logit $s_h$ that absorbs excess attention mass but contributes nothing to the value aggregation. We enable this sink in all softmax-attention layers.

\subsection{Lightning Attention}
Lightning attention is a representative linear-attention variant within the recurrent sequence mixer family introduced in Section~\ref{sec:pre:hybrid}. Compared with a full-attention layer, a Lightning layer introduces a small number of additional parameters. To keep the Lightning hybrid comparable with \emph{Full} and the SWA hybrids in total parameter count, we preserve the GQA configuration, layer count, and backbone hidden size, and only slightly reduce the FFN hidden size inside the Lightning layers. The resulting configuration is summarized in Table~\ref{tab:lightning_params}.

\begin{table}[t]
    \centering
    \small
    \setlength{\tabcolsep}{8pt}
    \renewcommand{\arraystretch}{1.15}
    \begin{tabular}{@{}lrrrr@{}}
        \toprule
        \multirow{2}{*}{Scale} & \multicolumn{2}{c}{FFN hidden} & \multicolumn{2}{c}{Per-layer params} \\
        \cmidrule(lr){2-3} \cmidrule(lr){4-5}
                       & Full      & Lightning   & Full              & Lightning          \\
        \midrule
        S1 &   960 &   920 & 1{,}499{,}136 & 1{,}502{,}208 \\
        S2 & 1{,}280 & 1{,}240 & 2{,}621{,}440 & 2{,}625{,}536 \\
        S3 & 1{,}600 & 1{,}560 & 4{,}055{,}040 & 4{,}060{,}160 \\
        S4 & 1{,}920 & 1{,}880 & 5{,}799{,}936 & 5{,}806{,}080 \\
        \bottomrule
    \end{tabular}
    \caption{Lightning configuration and per-layer parameter counts. Per-layer params report the parameter counts of a single layer; shared LayerNorms and embeddings are omitted.}
    \label{tab:lightning_params}
\end{table}

\begin{table}[t]
    \centering
    \small
    \setlength{\tabcolsep}{8pt}
    \renewcommand{\arraystretch}{1.15}
    \begin{tabular}{@{}lcrr@{}}
        \toprule
        \multirow{2}{*}{Scale} & \multirow{2}{*}{Head Dim} & \multicolumn{2}{c}{Per-layer params} \\
        \cmidrule(lr){3-4}
                       &    & Full          & GDN           \\
        \midrule
        S1 & 46 & 1{,}499{,}136 & 1{,}496{,}114 \\
        S2 & 46 & 2{,}621{,}440 & 2{,}627{,}634 \\
        S3 & 46 & 4{,}055{,}040 & 4{,}075{,}570 \\
        S4 & 46 & 5{,}799{,}936 & 5{,}839{,}922 \\
        \bottomrule
    \end{tabular}
    \caption{Gated DeltaNet configuration and per-layer parameter counts. ``Head Dim'' is the per-group key/value channel dimension $d_k = d_v$ (since \texttt{expand\_v} $= 1$); the Full backbone uses $d_k = d_v = 64$.}
    \label{tab:gdn_params}
\end{table}

\begin{table}[t]
    \centering
    \small
    \setlength{\tabcolsep}{8pt}
    \renewcommand{\arraystretch}{1.15}
    \begin{tabular}{@{}lcrr@{}}
        \toprule
        \multirow{2}{*}{Scale} & \multirow{2}{*}{State Dim} & \multicolumn{2}{c}{Per-layer params} \\
        \cmidrule(lr){3-4}
                       &     & Full          & Mamba-2       \\
        \midrule
        S1 &  16 & 1{,}499{,}136 & 1{,}576{,}098 \\
        S2 &  16 & 2{,}621{,}440 & 2{,}789{,}912 \\
        S3 &  16 & 4{,}055{,}040 & 4{,}349{,}470 \\
        S4 &  16 & 5{,}799{,}936 & 6{,}253{,}092 \\
        \bottomrule
    \end{tabular}
    \caption{Mamba-2 configuration and per-layer parameter counts. State Dim is the SSM state dimension $d_{\text{state}}$. Mamba-2 ends up slightly larger (5\%--8\%) than Full in order to retain sufficient state capacity.}
    \label{tab:mamba2_params}
\end{table}

\begin{table*}[t]
    \centering
    \small
    \setlength{\tabcolsep}{10pt}
    \renewcommand{\arraystretch}{1.15}
    \begin{tabular}{@{}lcccccc@{}}
        \toprule
        \textbf{Training tokens $D$} & $100N$ & $200N$ & $300N$ & $400N$ & $500N$ & $1000N$ \\
        \midrule
        \multicolumn{7}{@{}l}{\textit{C4 validation loss ($\downarrow$)}} \\
        \quad GDN w/ conv1d   & 4.336 & 4.163 & 4.088 & 4.053 & 4.014 & 3.929 \\
        \quad GDN w/o conv1d  & 4.368 & 4.179 & 4.106 & 4.072 & 4.028 & 3.942 \\
        \addlinespace
        \multicolumn{7}{@{}l}{\textit{LongPPL ($\downarrow$)}} \\
        \quad GDN w/ conv1d   & 80.79 & 19.23 & 15.91 & 13.31 & 12.85 & 11.36 \\
        \quad GDN w/o conv1d  & 91.35 & 24.96 & 16.44 & 13.56 & 12.30 & 11.01 \\
        \bottomrule
    \end{tabular}
    \caption{Ablation of the short 1D convolution in Gated DeltaNet at the S1 scale. The convolution consistently lowers C4 validation loss by a small margin throughout training. Its LongPPL advantage, however, exists only at small training budgets: the gap closes to within $0.5$ by $D{\ge}300N$ and reverses at $D{\ge}500N$.}
    \label{tab:gdn_conv_ablation}
\end{table*}

\subsection{Gated DeltaNet}

Gated DeltaNet (GDN) is a more elaborate recurrent sequence mixer that combines the gated update of GLA \citep{yang2024gated} with the delta-rule mechanism of DeltaNet \citep{schlag2021linear,yang2024parallelizing} (Section~\ref{sec:pre:hybrid}). The standard GDN implementation additionally includes a short 1D convolution on Q/K/V and a value-expansion factor (\texttt{expand\_v}) that widens the value dimension relative to the key dimension. To make the GDN hybrid comparable to \emph{Full} and the other hybrid variants under a matched parameter budget, we make two adjustments to this configuration.

\paragraph{Removing the short convolution.} The short 1D convolution on Q/K/V is an auxiliary mixing operator not common in Transformer-based models, and keeping it would conflate the effect of the recurrent sequence mixer itself with this auxiliary mechanism. We disable it in our main study, and verify with a small ablation at the S1 scale (Table~\ref{tab:gdn_conv_ablation}). The convolution consistently improves C4 validation loss by a small margin throughout training, but its $\mathrm{LongPPL}$ advantage exists only at small training budgets and vanishes once the budget is sufficient. Therefore, disabling the short convolution simplifies the architectural comparison without altering the long-context findings in the paper.

\paragraph{Keeping FFN width and tuning the state dimension.} At default settings, a GDN layer is heavier than a Full attention layer due to its data-dependent gating and recurrent-state projections. Shrinking the FFN to compensate would require an awkwardly narrow width, so we instead keep the FFN identical to the Full backbone and adjust only the state-related dimensions. We find that $\texttt{expand\_v}=1$ (i.e., $d_v = d_k$) consistently outperforms $d_v > d_k$ on validation loss, and therefore fix $\texttt{expand\_v}=1$ and pick the GDN head dimension so the per-layer parameter count matches \emph{Full}, giving $d_k = d_v = 46$ (Table~\ref{tab:gdn_params}).

\subsection{Mamba-2}
As with GDN, the standard Mamba-2 implementation includes a short 1D causal convolution before the SSM and an expansion factor $\texttt{expand}$ that widens the SSM hidden dimension. We apply the same strategy as for GDN: disable the convolution to isolate the recurrence, and adjust only the SSM-related dimensions while keeping the FFN unchanged.

A default Mamba-2 layer is also heavier than a Full attention layer, mainly due to the SSM projections ($\Delta_t$, $B$, $C$, $A$) together with the input/output projections. Mamba-2 provides a dedicated $\texttt{state\_dim}$ parameter that controls the SSM state size independently of the per-head channel width $\texttt{head\_dim}$, which we use to match the per-layer parameter count to \emph{Full}. Specifically, we set $\texttt{expand}=1$, keep $\texttt{head\_dim}=64$ to match Full's attention head dimension, and shrink $\texttt{state\_dim}$ to $16$, which still preserves sufficient state capacity (Table~\ref{tab:mamba2_params}).

\section{Training Details}
\label{app:training}
For the S1--S4 models, we share the same training hyperparameters (data, sequence length, learning-rate schedule, and batch size) across architectures, so that the scaling comparison is not confounded by optimization or data differences. All models are pretrained with a 16K sequence length, a $1{:}1$ mixture of long and short documents, and a Warmup-Stable-Decay (WSD) learning-rate schedule \citep{hu2024minicpm}. The stable and decay phases account for $90\%$ and $10\%$ of the total training tokens, respectively. During the decay phase, the learning rate is linearly annealed from the stable value to $1/10$ of it. For scaling-law fitting, we use checkpoints at $D/N \in \{100, 200, 300, 400, 500, 1000\}$ for S1--S4 (and $D/N \in \{100, 200\}$ for S5); each checkpoint corresponds to a complete WSD schedule ($90\%$ stable plus $10\%$ decay scaled to that $D/N$), not a mid-stable snapshot.

Table~\ref{tab:training_schedule} summarizes the concrete schedule: S1--S4 are trained to $D/N{=}1000$ while S5 is trained to $D/N{=}200$. The global batch size and stable learning rate at each scale were obtained from a hyperparameter sweep, and we report a configuration that consistently performs well on the Full baseline. For fairness, the same configuration is then shared by all hybrid variants at that scale.

The final row of Table~\ref{tab:training_schedule} records the long-context extension of the S5/$200N$ checkpoint used in Section~\ref{sec:discussion}: we continue training for $\approx 5B$ tokens ($4{,}769$ iters) at a 32K sequence length, with the LR linearly decayed from the S5 end LR to $0$ (no stable phase) and the RoPE base raised from $10^5$ to $5\times10^5$.

\begin{table*}[htbp]
    \centering
    \setlength{\tabcolsep}{4pt}
    \small
    \resizebox{\textwidth}{!}{%
    \begin{tabular}{lcccccccccc}
        \toprule
        Scale & $D/N$ & Global batch & Stable LR & End LR & Stable iters & Decay iters & Total iters & Seq.\ len.\ & RoPE base \\
        \midrule
        S1 & 1000 & 32 & $1.953\times10^{-3}$ & $1.953\times10^{-4}$ & 25{,}764  & 2{,}862  & 28{,}626  & 16K & $10^5$ \\
        S2 & 1000 & 16 & $9.766\times10^{-4}$ & $9.766\times10^{-5}$ & 111{,}938 & 12{,}438 & 124{,}376 & 16K & $10^5$ \\
        S3 & 1000 & 28 & $9.542\times10^{-4}$ & $9.542\times10^{-5}$ & 119{,}740 & 13{,}304 & 133{,}044 & 16K & $10^5$ \\
        S4 & 1000 & 64 & $9.766\times10^{-4}$ & $9.766\times10^{-5}$ & 94{,}318  & 10{,}480 & 104{,}798 & 16K & $10^5$ \\
        S5 &  200 & 64 & $9.766\times10^{-4}$ & $9.766\times10^{-5}$ & 81{,}900  & 9{,}100  & 91{,}000  & 16K & $10^5$ \\
        \midrule
        S5 + 32K ext.\ & --   & 32 & $9.766\times10^{-5}$ & $0$ & 0 & 4{,}769 & 4{,}769 & 32K & $5\times10^5$ \\
        \bottomrule
    \end{tabular}
    }
    \caption{Training schedule. ``$D/N$'' is the training budget; iters columns report actual training iters. The final row is the long-context extension used in Section~\ref{sec:discussion}. All runs use the Muon optimizer with weight decay $0.1$~\citep{jordan2024muon}, gradient clipping $1.0$.}
    \label{tab:training_schedule}
\end{table*}

\section{Mechanism Analysis Details}
\label{app:mechanism}
We conduct several experiments to analyze the mechanism of long-range retrieval in hybrid models, including probing, receptive-field constraints, gradient profiling, and retrieval-head tracing. Here, we provide implementation details for these experiments and explain why they support our conclusions that full attention dominates long-range retrieval and that efficient attention shapes long-context training dynamics by modulating the optimization pressure on full attention.

\subsection{Receptive-field Constraint Details}

This section gives implementation details for the inference-time receptive-field restriction experiment in Section~\ref{sec:analysis:ga-dominates} (Figure~\ref{fig:horizon}), where we limit either full attention or efficient attention to a receptive field of $H{\approx}2048$ tokens and measure the change in $\log(\mathrm{LongPPL})$.

\paragraph{Softmax attention (Full and SWA).}
We apply an exact 4D attention mask: for a query at position $i$, attention is allowed only to keys at positions in $[\,i-H,\,i\,]$ with $H=2048$. This gives a strict per-token receptive field.

\paragraph{Recurrent kernels (Lightning, Mamba-2, GDN).}
The same masking cannot be applied to the recurrent/SSM kernels. We instead use an overlapping-window approximation: the sequence is split into windows of $3072$ tokens with a $1024$-token stride; within each window, the recurrent state is reset to zero and rolled forward, and only the last $1024$ positions of the window are written to the output buffer. Concretely, for a retained block starting at position $s\ge 2048$, the computation window is $[s-2048,\,s+1024)$ and the copied-back interval is $[s,\,s+1024)$, so each token's recurrent state is built from $\approx 2049$ to $3072$ preceding tokens. The effective receptive field is therefore slightly looser than the strict $H{=}2048$ used for softmax attention, but is well within the same order of magnitude; we report this as the same ``$H{\approx}2048$'' condition in Section~\ref{sec:analysis:ga-dominates}.

\subsection{Layer-wise Probing Analysis}
\label{app:probing}
This section gives implementation details for the layer-wise probing experiment in Section~\ref{sec:analysis:ga-dominates} (Figure~\ref{fig:probing}). We probe the S4/$1000N$ checkpoints of seven models: \emph{Full}, \emph{SWA-128}, \emph{SWA-512}, \emph{SWA-2048}, \emph{Lightning}, \emph{Mamba-2}, and \emph{GDN}. The synthetic NIAH classification dataset contains 10{,}000 samples with a sequence length of 16K and eight candidate classes; its prompt format is illustrated in Figure~\ref{fig:probing_prompt_format}.

\begin{figure*}[htbp]
    \centering
    \begin{tcolorbox}[title=NIAH probing sample format, width=\textwidth]
    \textbf{Candidate values}: \texttt{31415920}, \texttt{31415921}, \ldots, \texttt{31415927}

    \medskip
    \textbf{Sample fields}:
    \begin{itemize}
        \item Key: \texttt{golden-crystal} \hfill sampled adjective--noun identifier
        \item Value: \texttt{31415923} \hfill one value from the eight candidates
        \item Label: $3$ \hfill index of the selected value
    \end{itemize}

    \medskip
    \textbf{Data}:
    \begin{quote}
    A special magic number is hidden within the following text. Make sure to memorize it. I will quiz you about the number afterwards.

    \textit{[repeated distractor passage]}

    One of the special magic numbers for \texttt{golden-crystal} is: \texttt{31415923}.

    \textit{[repeated distractor passage]}

    What is the special magic number for \texttt{golden-crystal} mentioned in the provided text? The special magic number for \texttt{golden-crystal} mentioned in the provided text is
    \end{quote}
    \end{tcolorbox}
    \caption{Data format for the NIAH classification dataset used in layer-wise probing. The probe predicts the label of the inserted magic number from the final query-token hidden state.}
    \label{fig:probing_prompt_format}
\end{figure*}

For each model and each sample, we run a forward pass with hidden-state output enabled and extract the hidden state of the final query token after every transformer layer. We train an independent logistic-regression probe for each layer, using an 80/20 train/test split with stratified labels and standardizing the hidden states before fitting; the multi-class implementation uses a one-vs-rest scheme. Table~\ref{tab:probing_classifier_ablation} shows that logistic regression gives the strongest layer-wise accuracy among the lightweight classifiers we test, so we use it as the primary probe.

Figure~\ref{fig:probing} visualizes the incremental layer contribution, i.e., the heatmap entries are $A_\ell-A_{\ell-1}$ where $A_\ell$ is the raw probing accuracy at layer $\ell$. Table~\ref{tab:probing_layerwise_results} reports the underlying raw layer-wise accuracies for all 18 layers.

Interestingly, probing accuracy typically peaks at intermediate layers and declines in deeper layers. This suggests that retrieval-related information becomes most linearly accessible in the middle layers, while later layers progressively mix and integrate these signals into higher-level semantic representations, making them less separable by lightweight classifiers. This observation is broadly consistent with prior findings that transformer representations evolve from surface and syntactic features in lower and middle layers toward more abstract semantic representations in deeper layers~\citep{jawahar-etal-2019-bert}.

\begin{table*}[t]
    \centering
    \scriptsize
    \setlength{\tabcolsep}{2pt}
    \resizebox{\textwidth}{!}{%
    \begin{tabular}{lcccccccccccccccccc}
        \toprule
        Classifier & L0 & L1 & L2 & L3 & L4 & L5 & L6 & L7 & L8 & L9 & L10 & L11 & L12 & L13 & L14 & L15 & L16 & L17 \\
        \midrule
        Logistic regression & 19.4 & 16.1 & 13.7 & 14.1 & 17.6 & 16.1 & 14.1 & 14.5 & 16.5 & 47.7 & 95.3 & 92.8 & 92.3 & 89.6 & 86.8 & 82.1 & 78.8 & 74.9 \\
        MLP & 22.1 & 13.4 & 11.8 & 12.6 & 12.0 & 11.8 & 12.3 & 11.2 & 11.7 & 28.9 & 91.2 & 87.1 & 84.8 & 79.7 & 68.8 & 64.8 & 63.0 & 54.9 \\
        Random forest & 30.6 & 15.6 & 13.9 & 13.1 & 14.0 & 12.1 & 12.3 & 11.8 & 12.3 & 11.8 & 18.4 & 18.8 & 16.4 & 15.8 & 14.8 & 14.4 & 14.4 & 13.1 \\
        kNN & 20.7 & 14.3 & 13.2 & 12.6 & 12.3 & 12.7 & 11.5 & 12.3 & 12.1 & 12.2 & 13.8 & 13.1 & 13.3 & 13.5 & 12.7 & 12.7 & 12.4 & 12.6 \\
        PCA+Naive Bayes & 15.6 & 12.6 & 12.7 & 12.9 & 13.6 & 14.1 & 12.2 & 11.3 & 11.1 & 16.6 & 65.5 & 57.9 & 55.1 & 51.0 & 39.0 & 30.8 & 28.7 & 28.2 \\
        \bottomrule
    \end{tabular}
    }
    \caption{Comparison of lightweight classifiers on the S4/$1000N$ Full model under the same NIAH probing task as Table~\ref{tab:probing_layerwise_results}; logistic regression gives the strongest layer-wise accuracy.}
    \label{tab:probing_classifier_ablation}
\end{table*}

\begin{table*}[t]
    \centering
    \scriptsize
    \setlength{\tabcolsep}{2pt}
    \resizebox{\textwidth}{!}{%
    \begin{tabular}{lcccccccccccccccccc}
        \toprule
        Model & L0 & L1 & L2 & L3 & L4 & L5 & L6 & L7 & L8 & L9 & L10 & L11 & L12 & L13 & L14 & L15 & L16 & L17 \\
        \midrule
        Full & 19.4 & 16.1 & 13.7 & 14.1 & 17.6 & 16.1 & 14.1 & 14.5 & 16.5 & 47.7 & 95.3 & 92.8 & 92.3 & 89.6 & 86.8 & 82.1 & 78.8 & 74.9 \\
        GDN & 19.9 & 25.1 & 21.1 & 25.4 & 28.1 & 26.8 & 28.2 & 32.7 & 27.1 & 74.4 & 63.8 & 60.0 & 58.1 & 77.5 & 77.0 & 67.5 & 64.2 & 55.6 \\
        Lightning & 12.1 & 12.2 & 12.4 & 11.9 & 12.7 & 11.6 & 12.8 & 23.5 & 23.0 & 67.0 & 64.5 & 89.1 & 80.2 & 82.2 & 78.4 & 72.0 & 68.8 & 63.4 \\
        Mamba-2 & 12.1 & 13.7 & 12.1 & 14.2 & 14.9 & 14.0 & 12.8 & 12.5 & 13.8 & 16.7 & 15.1 & 61.7 & 53.6 & 78.2 & 69.7 & 57.5 & 51.8 & 35.8 \\
        SWA-128 & 12.3 & 11.5 & 12.5 & 12.0 & 12.4 & 14.3 & 12.6 & 39.2 & 33.1 & 76.6 & 61.5 & 75.8 & 69.1 & 85.0 & 80.2 & 77.5 & 73.7 & 67.7 \\
        SWA-512 & 11.6 & 12.7 & 13.5 & 11.7 & 12.6 & 11.8 & 12.2 & 22.6 & 28.3 & 86.2 & 78.6 & 87.3 & 81.4 & 75.6 & 69.9 & 65.7 & 63.5 & 60.0 \\
        SWA-2048 & 12.4 & 12.5 & 12.7 & 13.2 & 15.8 & 28.5 & 29.0 & 34.2 & 32.6 & 69.0 & 66.2 & 72.8 & 64.6 & 61.4 & 57.0 & 53.2 & 50.0 & 45.5 \\
        \bottomrule
    \end{tabular}
    }
    \caption{Layer-wise logistic-regression probing accuracy on the S4/$1000N$ NIAH classification task.}
    \label{tab:probing_layerwise_results}
\end{table*}

\subsection{Gradient Profiling}
\label{app:gradient_profiling}

Gradient profiling uses the input-gradient norm of a logit-based scalar output as a proxy for the long-range training signal that a historical token provides for next-token prediction. We give a short derivation linking this proxy to (i)~local sensitivity of the model's prediction, (ii)~gradients on retrieval-head Q/K parameters, and (iii)~conditional dependency in the data, and we use it to read Figure~\ref{fig:gradient_influence}.

Let $x_{1:T} \sim \mathcal{D}$ be a token sequence sampled from the pretraining distribution, $e_i \in \mathbb{R}^{d_{\mathrm{model}}}$ the input embedding of $x_i$, and $z^{(t)}(x) \in \mathbb{R}^{|\mathcal{V}|}$ the logit vector produced by $p_\theta$ at position $t$. Following \citet{li2016visualizing}, we summarize the model's prediction near the end of the context by the scalar
\[
    s(x)
    \;=\;
    \sum_{v \in \mathcal{V}}
    \frac{1}{N_\tau}\sum_{t \in \tau} z_v^{(t)}(x),
\]
where $\tau$ is the last $N_\tau = 20$ positions, and report the average input-gradient norm at distance $d = T - i$,
\[
    G(d)
    \;=\;
    \mathbb{E}_{x \sim \mathcal{D}}\!\left[
    \left\|\partial s(x) / \partial e_{T-d}\right\|_2
    \right].
\]

\paragraph{(1) Local sensitivity.}
A first-order Taylor expansion of $s$ in $e_i$, followed by Cauchy--Schwarz, gives for any perturbation $\Delta e_i$, up to second-order terms in $\|\Delta e_i\|$,
\[
    \big| s(e_i + \Delta e_i) - s(e_i) \big|
    \;\le\;
    \|\partial s / \partial e_i\|_2 \cdot \|\Delta e_i\|_2.
\]
So $\|\partial s / \partial e_i\|_2$ tightly bounds the first-order change of $s$ under infinitesimal perturbations of $e_i$.

\paragraph{(2) Connection to retrieval-head gradients.}
By chain rule, $\partial s / \partial e_i$ decomposes into contributions from all computational paths that route information from position $i$ into the last $N_\tau$ positions. For a single retrieval head with attention weights $a_{t,j}$ and per-position output $o_t = \sum_j a_{t,j} v_j$ (with $v_j = V e_j$), a direct softmax computation gives
\[
    \frac{\partial s}{\partial \mathrm{score}_{t,i}}
    \;=\;
    a_{t,i}\,(v_i - o_t)^{\!\top}\frac{\partial s}{\partial o_t},
\]
so the head's Q/K gradient at the entry $(t,i)$ shares the multiplicative factor $a_{t,i}\,\partial s/\partial o_t$. The same factor also appears in the value-path contribution to $\partial s/\partial e_i$, via $\partial s/\partial v_i = \sum_{t\in\tau} a_{t,i}\,\partial s/\partial o_t$. Hence, absent fine-tuned path cancellation, a small $\|\partial s/\partial e_i\|_2$ implies that the Q/K update strengthening retrieval at distance $d$ is correspondingly weak, and we read $G(d)$ as a per-sample upper-bound proxy on this training signal.

\paragraph{(3) Connection to data dependency.}
If the data satisfies the conditional independence $y_t \perp x_i \mid x_{i+1:t}$ for every $t \in \tau$, then a sufficiently trained $p_\theta$ inherits the same independence in its predictive distribution, and the gradient vanishes:
\[
    \begin{aligned}
        y_t \perp x_i \mid x_{i+1:t}
        &\;\Longrightarrow\;
        p_\theta(\cdot \mid x_{1:t}) \approx p_\theta(\cdot \mid x_{i+1:t}) \\
        &\;\Longrightarrow\;
        \partial s/\partial e_i \approx 0.
    \end{aligned}
\]
Conversely, a genuine conditional dependency at distance $d$ forces $\partial s/\partial e_i$ to be nonzero on average. Crucially, $y_t \perp x_i \mid x_{i+1:t}$ is a property of the \emph{data distribution}, so the dependency profile reflected by $G(d)$ transfers across models trained on similar corpora; this justifies using Llama-3.1-8B as a proxy for the dependency signal seen by our hybrid models.\footnote{Strictly, conditional independence constrains logits only up to a global additive constant (softmax is invariant under such shifts); in the standard parameterization $z_v^{(t)} = w_v^{\!\top} h^{(t)}$, this common mode carries no independent training signal.}

Combining (1)--(3), for a sufficiently trained $p_\theta$, a small $G(d)$ jointly indicates local insensitivity of $s$ to $e_{T-d}$, weak Q/K updates that would strengthen retrieval at distance $d$, and weak conditional dependency at distance $d$ in the data.

\paragraph{The flat baseline.}
Even when $x_i$ is conditionally uninformative, $G(d)$ does not reach zero in practice; instead, it decays to a flat baseline. Three sources contribute to this irreducible level: (a) finite-precision backward arithmetic, (b) finite-capacity $p_\theta$ that is not exactly Bayes-optimal, and (c) coarse topic/style/domain signals that distant tokens still carry. Formally, even with a mean-zero per-sample gradient, Jensen's inequality gives
\[
    G(d) \;=\; \mathbb{E}\!\left[\|\partial s/\partial e_i\|_2\right]
    \;\ge\;
    \|\mathbb{E}[\partial s/\partial e_i]\|_2,
\]
so $G(d)$ remains strictly positive whenever the per-sample gradient is non-degenerate. We therefore estimate the baseline at a distance where Figure~\ref{fig:gradient_influence} has visibly flattened,
\[
    G_{\mathrm{base}}
    \;:=\;
    G_{\mathrm{PG19}}(d = 4096),
\]
shown as the dashed reference line in Figure~\ref{fig:gradient_influence}, and treat $G(d) \lesssim G_{\mathrm{base}}$ as effectively no usable retrieval signal at distance $d$. Figure~\ref{fig:gradient_influence} then becomes a quantitative map of the distance ranges that contribute training signal during pretraining, which directly supports the Large-Window Laziness argument in Section~\ref{sec:analysis:prior}: a SWA window already covering the range where $G(d) \gg G_{\mathrm{base}}$ absorbs most of the dependency-driven training signal before it can propagate to full-attention retrieval heads.

\subsection{Retrieval-Head Tracing}
\label{sec:retrieval}
This section gives implementation details for the retrieval-head tracing experiment in Section~\ref{sec:analysis:prior} (Figure~\ref{fig:tracing}) and more analysis around the formation of the retrieval head in hybrid architectures.

\paragraph{NIAH probe and head score.}
We construct an NIAH probe where a unique ``needle'' string is hidden in a long context and the prompt ends with a question whose answer is the needle. Running the S4/$200N$ checkpoint of each hybrid on this prompt, we read the per-head attention from the last input position $q$ (the query) to all keys, and score each head $(\ell,h)$ by the attention mass it places on the needle tokens, averaged over NIAH samples:
\[
    \overline{\mathrm{score}}_{\ell,h}
    \;=\;
    \frac{1}{|\mathcal{S}|}\sum_{x \in \mathcal{S}}
    \sum_{j \in \mathcal{N}(x)} a^{(\ell,h)}_{q,\,j}(x),
\]
where $a^{(\ell,h)}_{q,j}(x)$ is the attention weight from $q$ to key $j$ in head $(\ell,h)$ for sample $x$, $\mathcal{N}(x)$ is the set of needle token positions, and $\mathcal{S}$ is the NIAH evaluation set. A high $\overline{\mathrm{score}}_{\ell,h}$ means the head consistently routes the query's attention back to the needle---the canonical retrieval-head signature.

\begin{figure*}[!t]
    \centering
    \includegraphics[width=0.92\textwidth]{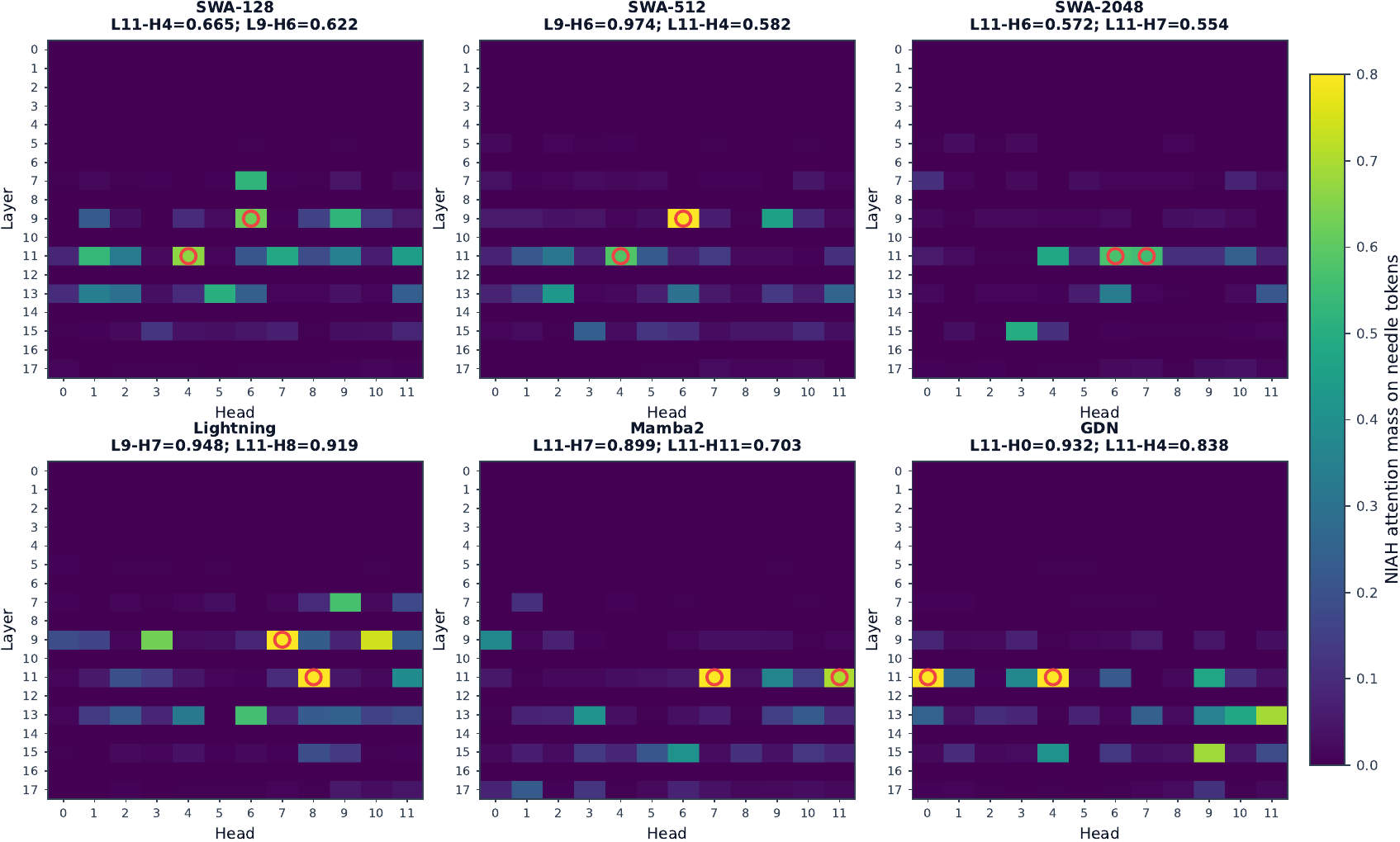}
    \caption{Per-head NIAH attention-mass scores $\overline{\mathrm{score}}_{\ell,h}$ for the six S4/$200N$ hybrid models. Red circles mark the selected top-2 retrieval heads in each model.}
    \label{fig:retrieval_head_heatmaps}
\end{figure*}

\paragraph{Head selection.}
Each cell of Figure~\ref{fig:retrieval_head_heatmaps} reports $\overline{\mathrm{score}}_{\ell,h}$ for one (layer, head), for the six traced hybrid models: \emph{SWA-128}, \emph{SWA-512}, \emph{SWA-2048}, \emph{Lightning}, \emph{Mamba-2}, and \emph{GDN}. We restrict the search to full-attention layers, since our analysis targets long-range retrieval formed there, and select the Top-2 heads per model (red circles) as the retrieval-head set used by the tracing diagnostics in Section~\ref{sec:analysis:prior}; lower-ranked heads have noisier retrieval signatures and would dilute these diagnostics.

Figure~\ref{fig:retrieval_head_heatmaps} also reveals that \emph{SWA-2048} has noticeably fewer high-response heads in its full-attention layers than the other hybrids, consistent with the \emph{Large-Window Laziness} hypothesis.

\begin{figure*}[!t]
    \centering
    \includegraphics[width=0.92\linewidth]{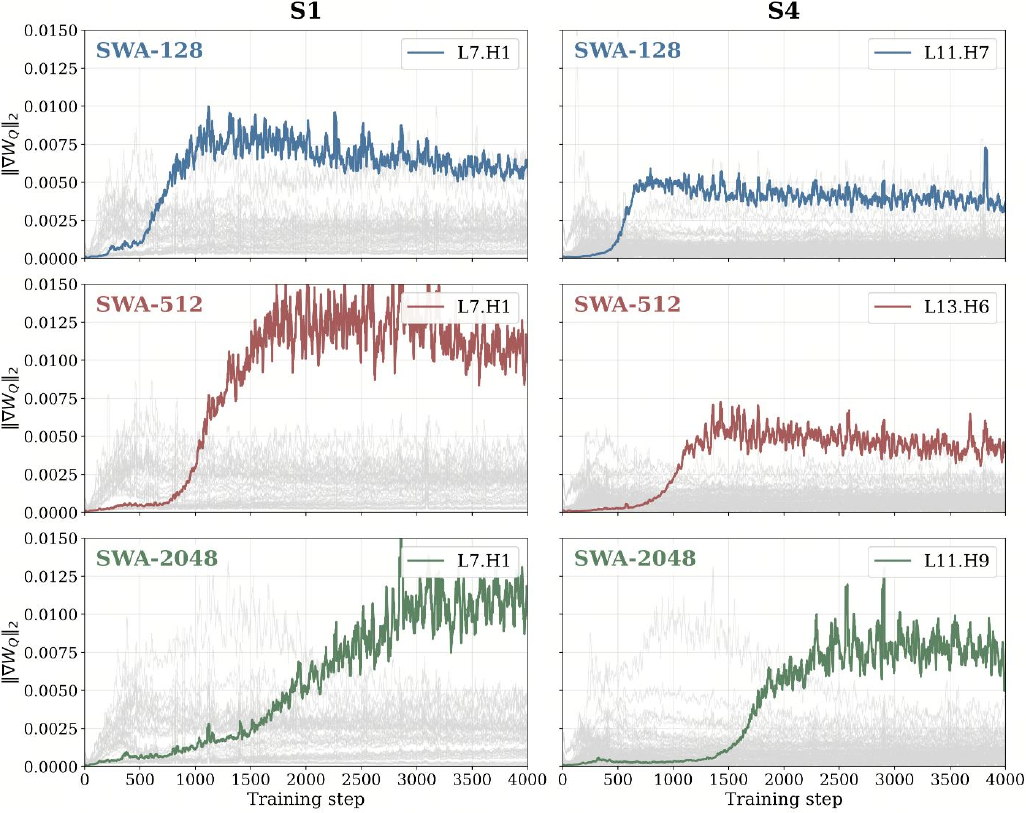}
    \caption{Smaller sliding-window attention activates retrieval-head training earlier. The figure shows the evolution of the Frobenius norm of the gradient on the $Q$ projections of retrieval heads during training for \emph{SWA-128}, \emph{SWA-512}, and \emph{SWA-2048} under both S1 and S4 model scales.}
    \label{fig:retrieval_head_grad_norm}
\end{figure*}

\paragraph{Training Gradient.}
To trace the training dynamics of retrieval heads, we train SWA hybrid models with different window sizes and track their gradient norms throughout training. Following the setup of our scaling experiments in Appendix~\ref{app:training}, we train S1- and S4-scale models from scratch using a constant learning rate for $4000$ steps, with an initial $100$-step warmup phase. During training, we record the gradients of the $Q$ projection slices for all heads, and use the final checkpoint to identify the Top-1 retrieval head. We then compare the evolution of the gradient norm of this retrieval head throughout training.

Figure~\ref{fig:retrieval_head_grad_norm} shows the evolution of the Frobenius norm of the loss gradient with respect to the $Q$ projection weights of the retrieval head during training:
\[
\|\nabla W\|_F
=
\left(
\sum_{i,j}
\left(
\frac{\partial \mathcal{L}}{\partial W_{ij}}
\right)^2
\right)^{1/2}.
\]

We can clearly observe that smaller sliding windows allocate gradient mass to retrieval heads much earlier, whereas larger sliding windows substantially delay the training of retrieval heads. For example, the retrieval head in \emph{SWA-2048} does not begin to receive effective training until roughly $1500$ steps into training. The light gray curves in the figure represent the evolution of $\|\nabla W_Q\|_F$ for the other heads in the same model. Notably, for \emph{SWA-2048}, these other heads do not exhibit the same delayed-activation behavior.

\section{Benchmark Evaluation}
\label{sec:benchmark_evaluation}
Table~\ref{tab:downstream_verification} in the main paper reports only the aggregated scores. Here we provide the per-task results for the same configurations: \emph{Full}, \emph{SWA-128}, and \emph{SWA-128-NoPE}, each at S4 ($0.22B$) and S5 ($0.66B$) trained under ${\approx}100B$ tokens. The 16K-context results use these ${\approx}100B$-token checkpoints directly; the 32K-context results use the S5 checkpoint after an additional $5B$-token long-context extension at a 32K sequence length.

\paragraph{Benchmarks.}
For long-context evaluation, we use RULER \citep{hsieh2024ruler} and LongBench \citep{bai2024longbench}; for each RULER sub-task, we generate $200$ test instances and report task accuracy averaged over them. For short-context evaluation we use 19 standard benchmarks covering knowledge, commonsense reasoning, reading comprehension and natural language inference: MMLU \citep{hendrycks2021measuring}, C-Eval \citep{huang2023ceval}, CMMLU \citep{li2024cmmlu}, HellaSwag \citep{zellers2019hellaswag}, PIQA \citep{bisk2020piqa}, ARC-Easy and ARC-Challenge \citep{clark2018think}, WinoGrande \citep{sakaguchi2021winogrande}, OpenBookQA \citep{mihaylov2018can}, CommonsenseQA \citep{talmor2019commonsenseqa}, SIQA \citep{sap2019social}, StoryCloze \citep{mostafazadeh2016corpus}, RACE-middle and RACE-high \citep{lai2017race}, COPA \citep{roemmele2011choice}, RTE \citep{wang2019superglue}, CB \citep{demarneffe2019commitmentbank}, WiC \citep{pilehvar2019wic}, and MultiRC \citep{khashabi2018looking}.

\paragraph{Evaluation protocol.}
All evaluations use deterministic (greedy) decoding to eliminate sampling variance. For long-context tasks, we follow the task-specific reference-based metrics of RULER and LongBench. For short-context multiple-choice tasks, we score each candidate option by its log-likelihood (length-normalized perplexity) under the model and select the option with the highest score; this likelihood-based protocol better reflects the underlying capability of base models, which do not yet have the instruction-following ability needed for direct answer generation.

\paragraph{Per-task results.}
The detailed RULER-16K and LongBench scores are shown in Tables~\ref{tab:ruler_detailed} and~\ref{tab:longbench_detailed}; per-task short-context scores are reported in Table~\ref{tab:shortcontext_detailed}.

\section{Statement on the AI Usage}
During the writing and revision of this paper, the authors used large language models only as auxiliary tools for improving grammar, wording, sentence structure, clarity, and readability. These tools were not involved in the core academic work of this study, including formulating research questions, designing experiments, processing data, analyzing results, or drawing conclusions.

All LLM-assisted edits were carefully reviewed, judged, and revised by the authors. The authors take full responsibility for the authenticity, originality, accuracy, and completeness of the final manuscript.

\begin{table*}[t]
    \centering
    \small
    \setlength{\tabcolsep}{2pt}
    \renewcommand{\arraystretch}{1}
    \begin{tabular}{@{}lccccccccc@{}}
        \toprule
        \textbf{Task}
        & \multicolumn{3}{c}{\textbf{S4 / ${\approx}100B$ / $16K$}}
        & \multicolumn{3}{c}{\textbf{S5 / ${\approx}100B$ / $16K$}}
        & \multicolumn{3}{c}{\textbf{S5 / ${\approx}100B$ / $32K$}} \\
        \cmidrule(lr){2-4} \cmidrule(lr){5-7} \cmidrule(lr){8-10}
        & Full & SWA-128 & SWA-128-NoPE
        & Full & SWA-128 & SWA-128-NoPE
        & Full & SWA-128 & SWA-128-NoPE \\
        \midrule
        \rowcolor{black!4}\multicolumn{10}{@{}l}{\textit{NIAH: single key}} \\
        niah\_s1  & 81.00 & 97.00 & \textbf{100.00} & 99.50 & \textbf{100.00} & \textbf{100.00} & \textbf{100.00} & \textbf{100.00} & \textbf{100.00} \\
        niah\_s2  & 93.50 & 95.50 & \textbf{100.00} & 97.00 & 97.50 & \textbf{100.00} & \textbf{100.00} & 99.50 & 97.00 \\
        niah\_s3  & 27.50 & 41.00 & \textbf{66.50} & 65.00 & 64.00 & \textbf{66.50} & 54.00 & 50.50 & \textbf{63.50} \\
        \addlinespace
        \rowcolor{black!4}\multicolumn{10}{@{}l}{\textit{NIAH: multi key}} \\
        niah\_mk1 & 30.00 & 75.50 & \textbf{88.00} & \textbf{95.50} & 83.50 & 83.50 & \textbf{96.50} & 76.00 & 77.50 \\
        niah\_mk2 & 18.00 & 27.00 & \textbf{59.50} & 87.50 & 79.00 & \textbf{92.00} & \textbf{85.00} & 72.00 & 72.00 \\
        niah\_mk3 & 1.50 & 3.00 & \textbf{27.00} & 19.50 & 17.00 & \textbf{48.50} & 5.50 & 17.00 & \textbf{30.50} \\
        \addlinespace
        \rowcolor{black!4}\multicolumn{10}{@{}l}{\textit{NIAH: multi value / multi query}} \\
        niah\_mv  & 15.25 & 28.38 & \textbf{46.50} & 36.25 & 44.12 & \textbf{85.50} & 27.88 & 27.12 & \textbf{54.62} \\
        niah\_mq  & 20.88 & 29.25 & \textbf{55.00} & 36.88 & 42.12 & \textbf{82.50} & 32.00 & 39.25 & \textbf{68.25} \\
        \addlinespace
        \rowcolor{black!4}\multicolumn{10}{@{}l}{\textit{Variable tracking / aggregation}} \\
        vt        & \textbf{1.50} & 0.50 & 0.70 & 3.40 & \textbf{4.70} & 0.90 & \textbf{8.20} & 6.70 & 3.00 \\
        cwe       & \textbf{4.90} & 1.30 & 2.00 & 0.50 & \textbf{7.20} & 4.10 & 0.85 & \textbf{3.45} & 2.05 \\
        fwe       & 12.67 & \textbf{36.83} & 14.17 & \textbf{40.17} & 35.00 & 0.50 & \textbf{44.33} & 37.17 & 14.33 \\
        \addlinespace
        \rowcolor{black!4}\multicolumn{10}{@{}l}{\textit{QA}} \\
        qa\_1     & 14.00 & \textbf{16.00} & 15.50 & \textbf{20.50} & 18.00 & 15.50 & 8.00 & 8.50 & \textbf{17.00} \\
        qa\_2     & 5.50 & \textbf{8.00} & 7.50 & \textbf{11.50} & 7.50 & 8.00 & 8.50 & 7.00 & \textbf{11.00} \\
        \midrule
        \rowcolor{blue!3}\textit{NIAH average (8)}
        & 35.95 & 49.58 & \textbf{67.81}
        & 67.14 & 65.91 & \textbf{82.31}
        & 62.61 & 60.17 & \textbf{70.42} \\
        \rowcolor{blue!3}\textbf{Total average (13)}
        & 25.09 & 35.33 & \textbf{44.80}
        & 47.17 & 46.13 & \textbf{52.88}
        & 43.90 & 41.86 & \textbf{46.98} \\
        \bottomrule
    \end{tabular}
    \caption{Per-task results on RULER. ``NIAH average (8)'' is the average over the eight NIAH-style tasks; ``Total average (13)'' is the average over all $13$ RULER tasks.}
    \label{tab:ruler_detailed}
\end{table*}

\begin{table*}[t]
    \centering
    \small
    \setlength{\tabcolsep}{2pt}
    \renewcommand{\arraystretch}{1}
    \begin{tabular}{@{}lccccccccc@{}}
        \toprule
        \textbf{Task}
        & \multicolumn{3}{c}{\textbf{S4 / ${\approx}100B$ / $16K$}}
        & \multicolumn{3}{c}{\textbf{S5 / ${\approx}100B$ / $16K$}}
        & \multicolumn{3}{c}{\textbf{S5 / ${\approx}100B$ / $32K$}} \\
        \cmidrule(lr){2-4} \cmidrule(lr){5-7} \cmidrule(lr){8-10}
        & Full & SWA-128 & SWA-128-NoPE
        & Full & SWA-128 & SWA-128-NoPE
        & Full & SWA-128 & SWA-128-NoPE \\
        \midrule
        \rowcolor{black!4}\multicolumn{10}{@{}l}{\textit{Single-document QA}} \\
        narrativeqa      &  \textbf{2.82} &  2.52 &  2.53 &  2.72 &  2.93 &  \textbf{3.10} & 2.74 & 2.87 & \textbf{2.88} \\
        qasper           & 14.07 & \textbf{16.46} & 14.39 & 18.72 & 17.09 & \textbf{19.07} & 19.32 & 18.30 & \textbf{19.78} \\
        multifieldqa\_en & 16.06 & 17.69 & \textbf{17.72} & 19.09 & 19.70 & \textbf{21.01} & 20.26 & 20.39 & \textbf{21.39} \\
        multifieldqa\_zh & 13.19 & 13.06 & \textbf{13.35} & 17.17 & 16.08 & \textbf{17.60} & \textbf{17.59} & 15.26 & 15.74 \\
        \addlinespace
        \rowcolor{black!4}\multicolumn{10}{@{}l}{\textit{Multi-document QA}} \\
        hotpotqa         &  \textbf{6.70} &  6.28 &  6.28 &  7.34 &  \textbf{8.03} &  7.76 & 7.99 & 8.56 & \textbf{8.94} \\
        2wikimqa         &  \textbf{8.36} &  8.12 &  8.17 &  8.69 &  8.18 &  \textbf{9.27} & 8.58 & 8.44 & \textbf{9.77} \\
        musique          &  3.64 &  3.13 &  \textbf{3.78} &  3.78 &  4.05 &  \textbf{5.37} & 4.52 & 4.37 & \textbf{5.46} \\
        dureader         & 18.09 & \textbf{19.95} & 18.68 & 25.15 & 22.49 & \textbf{26.43} & 23.26 & 23.21 & \textbf{25.37} \\
        \addlinespace
        \rowcolor{black!4}\multicolumn{10}{@{}l}{\textit{Summarization}} \\
        gov\_report      & 15.33 & 20.60 & \textbf{26.76} & 24.46 & \textbf{24.48} & 23.40 & 26.15 & \textbf{29.54} & 25.68 \\
        qmsum            & 14.99 & \textbf{17.61} & 15.90 & \textbf{19.15} & 15.36 & 18.89 & \textbf{19.09} & 17.66 & 17.87 \\
        multi\_news      & 17.74 & 19.72 & \textbf{25.69} & 17.41 & 22.42 & \textbf{23.86} & 21.96 & 25.83 & \textbf{26.21} \\
        vcsum            &  0.90 &  5.65 &  \textbf{7.51} &  4.34 &  \textbf{5.18} &  4.39 & 2.14 & \textbf{6.54} & 5.04 \\
        \addlinespace
        \rowcolor{black!4}\multicolumn{10}{@{}l}{\textit{Few-shot learning}} \\
        trec             & \textbf{71.50} & 62.50 & 67.00 & 69.00 & 66.00 & \textbf{71.00} & 65.50 & 66.00 & \textbf{71.50} \\
        triviaqa         &  4.15 & \textbf{13.08} &  0.50 &  0.00 &  \textbf{3.03} &  0.50 & 0.00 & \textbf{0.50} & \textbf{0.50} \\
        samsum           & 12.74 & \textbf{18.06} &  9.39 & 27.83 & 18.34 & \textbf{30.35} & 28.34 & 18.12 & \textbf{29.76} \\
        lsht             &  6.00 &  6.50 & \textbf{12.00} & 15.50 & 16.25 & \textbf{21.00} & 21.00 & 23.50 & \textbf{24.25} \\
        \addlinespace
        \rowcolor{black!4}\multicolumn{10}{@{}l}{\textit{Synthetic}} \\
        passage\_count            &  \textbf{1.98} &  0.33 &  0.23 &  0.97 &  1.17 &  \textbf{3.13} & \textbf{2.35} & 0.62 & 0.96 \\
        passage\_retrieval\_en    &  3.83 &  3.71 &  \textbf{4.01} &  \textbf{4.17} &  3.67 &  3.83 & 3.54 & \textbf{4.79} & 3.88 \\
        passage\_retrieval\_zh    &  3.59 &  4.22 &  \textbf{4.53} &  3.85 &  \textbf{5.08} &  3.97 & 3.89 & \textbf{4.67} & 3.98 \\
        \addlinespace
        \rowcolor{black!4}\multicolumn{10}{@{}l}{\textit{Code completion}} \\
        lcc              & 43.88 & 38.11 & \textbf{45.79} & \textbf{48.70} & 44.39 & 41.27 & \textbf{49.49} & 41.22 & 43.52 \\
        repobench-p      & 37.33 & 36.16 & \textbf{40.78} & \textbf{49.24} & 44.06 & 44.14 & \textbf{49.88} & 43.84 & 46.09 \\
        \midrule
        \rowcolor{blue!3}\textbf{Average (21)}
        & 15.09 & 15.88 & \textbf{16.43}
        & 18.44 & 17.52 & \textbf{19.02}
        & 18.93 & 18.30 & \textbf{19.46} \\
        \bottomrule
    \end{tabular}
    \caption{Per-task results on LongBench, using the task-specific reference-based metrics from the official LongBench scripts. The bottom row averages all $21$ tasks and matches the LongBench column in Table~\ref{tab:downstream_verification}.}
    \label{tab:longbench_detailed}
\end{table*}

\begin{table*}[t]
    \centering
    \small
    \setlength{\tabcolsep}{4pt}
    \renewcommand{\arraystretch}{1.3}
    \begin{tabular}{@{}>{\raggedright\arraybackslash}p{0.18\linewidth}*{6}{>{\centering\arraybackslash}p{\dimexpr0.136667\linewidth-8pt\relax}}@{}}
        \toprule
        \textbf{Benchmark} & \multicolumn{3}{c}{\textbf{S4 / ${\approx}100B$}} & \multicolumn{3}{c}{\textbf{S5 / ${\approx}100B$}} \\
        \cmidrule(lr){2-4} \cmidrule(lr){5-7}
                         & Full  & SWA-128 & SWA-128-NoPE & Full  & SWA-128 & SWA-128-NoPE \\
        \midrule
        \rowcolor{black!4}\multicolumn{7}{@{}l}{\textit{Comprehensive knowledge}} \\
        MMLU              & \textbf{26.35} & 24.10 & 25.45 & 29.71 & 30.06 & \textbf{30.84} \\
        C-Eval            & \textbf{27.23} & 24.92 & 24.20 & 26.82 & 27.74 & \textbf{28.47} \\
        CMMLU             & \textbf{25.63} & 25.18 & 25.12 & 26.20 & \textbf{29.46} & 27.48 \\
        \addlinespace
        \rowcolor{black!4}\multicolumn{7}{@{}l}{\textit{Commonsense and completion}} \\
        HellaSwag         & 30.63 & \textbf{30.97} & 30.20 & 38.21 & \textbf{38.68} & 38.17 \\
        PIQA              & \textbf{64.09} & 61.92 & 61.92 & 65.83 & 65.89 & \textbf{66.21} \\
        ARC-Easy          & 39.51 & 39.15 & \textbf{40.04} & 43.03 & \textbf{44.80} & 42.15 \\
        ARC-Challenge     & 23.73 & 25.76 & \textbf{28.47} & \textbf{31.53} & 30.17 & 28.81 \\
        WinoGrande        & 52.17 & \textbf{53.91} & 52.80 & \textbf{54.38} & 53.99 & 53.43 \\
        OpenBookQA        & \textbf{27.60} & 26.60 & 27.40 & 25.00 & 24.80 & \textbf{27.60} \\
        CommonsenseQA     & 19.25 & 19.25 & \textbf{19.82} & 21.62 & \textbf{24.90} & 22.77 \\
        SIQA              & 38.08 & \textbf{39.41} & 38.08 & 40.63 & \textbf{41.50} & 40.58 \\
        StoryCloze        & \textbf{56.97} & 56.87 & 56.60 & 61.73 & 61.57 & \textbf{62.27} \\
        \addlinespace
        \rowcolor{black!4}\multicolumn{7}{@{}l}{\textit{Reading and entailment}} \\
        RACE-middle       & \textbf{25.07} & 21.87 & 23.12 & 27.51 & 30.15 & \textbf{34.75} \\
        RACE-high         & \textbf{25.64} & 21.13 & 21.70 & 28.16 & 29.96 & \textbf{30.62} \\
        COPA              & 51.00 & \textbf{54.00} & 49.00 & \textbf{58.00} & 56.00 & 57.00 \\
        RTE               & 48.74 & \textbf{53.43} & 53.07 & \textbf{52.71} & 51.62 & 50.54 \\
        CB                & \textbf{50.00} & \textbf{50.00} & \textbf{50.00} & 44.64 & \textbf{50.00} & \textbf{50.00} \\
        WiC               & \textbf{50.00} & \textbf{50.00} & \textbf{50.00} & \textbf{50.00} & \textbf{50.00} & \textbf{50.00} \\
        MultiRC           & 42.82 & \textbf{44.08} & 42.80 & 43.09 & \textbf{43.67} & 43.34 \\
        \midrule
        \rowcolor{blue!3}\textbf{Average (19)} & \textbf{38.13} & 38.03 & 37.88 & 40.46 & 41.31 & \textbf{41.32} \\
        \bottomrule
    \end{tabular}
    \caption{Per-task results on the 19 short-context benchmarks; bottom row averages all 19 tasks and matches the ShortAvg column in Table~\ref{tab:downstream_verification}. MMLU, C-Eval, and CMMLU report macro averages over their sub-tasks; the remaining rows report individual benchmark accuracies. All scores are obtained with deterministic decoding and option selection by length-normalized log-likelihood (higher is better).}
    \label{tab:shortcontext_detailed}
\end{table*}